\documentclass[letterpaper, 10 pt, conference]{ieeeconf}  %

\IEEEoverridecommandlockouts                              %

\usepackage[dvipsnames, table]{xcolor}
\usepackage{amsmath}
\usepackage{breqn}
\usepackage{float}
\usepackage{subcaption, graphicx}
\definecolor{red}{RGB}{255, 0, 0}   %
\definecolor{blue}{RGB}{0, 0, 255}   %
\definecolor{orange}{RGB}{255, 77, 0}   %
\definecolor{green}{RGB}{0, 128, 0}   %
\definecolor{purple}{RGB}{160, 32, 240}   %
\definecolor{lightblue}{RGB}{52, 155, 235}   %
\definecolor{darkmagenta}{RGB}{204, 51, 139}

\usepackage{todonotes}
\usepackage{outlines}
\usepackage[font=footnotesize]{caption}
\usepackage{titlesec}
\AtBeginDocument{
  \setlength\abovedisplayskip{0pt}
  \setlength\belowdisplayskip{0pt}
}

\usepackage{multirow}
\usepackage{hyperref}
\usepackage{cleveref}

\usepackage{booktabs}
\usepackage{xspace}
\usepackage{array}
\usepackage{tabularx}
\usepackage{pifont}%
\newcommand{\cmark}{\ding{51}}%
\newcommand{\xmark}{\ding{55}}%
\usepackage{makecell}

\usepackage{fontawesome}
\usepackage{scalerel}
\usepackage{rotating}

\usepackage[space]{cite}

\usepackage{pgf}
\definecolor{high}{HTML}{228B22}
\definecolor{low}{HTML}{FFFFFF}

\newcommand{\ourName}{AO-Grasp}
\newcommand{\ourData}{\ourName{} Dataset}
\newcommand{\ourNameFull}{Articulated Object Grasp Generation}

\newcommand{\ourModel}{AO-Grasp Model}
\newcommand{\modelPointscore}{Actionable Grasp Point Predictor}

\newcommand{\scoreName}{grasp-likelihood}

\def\trash{\scalerel*{\includegraphics{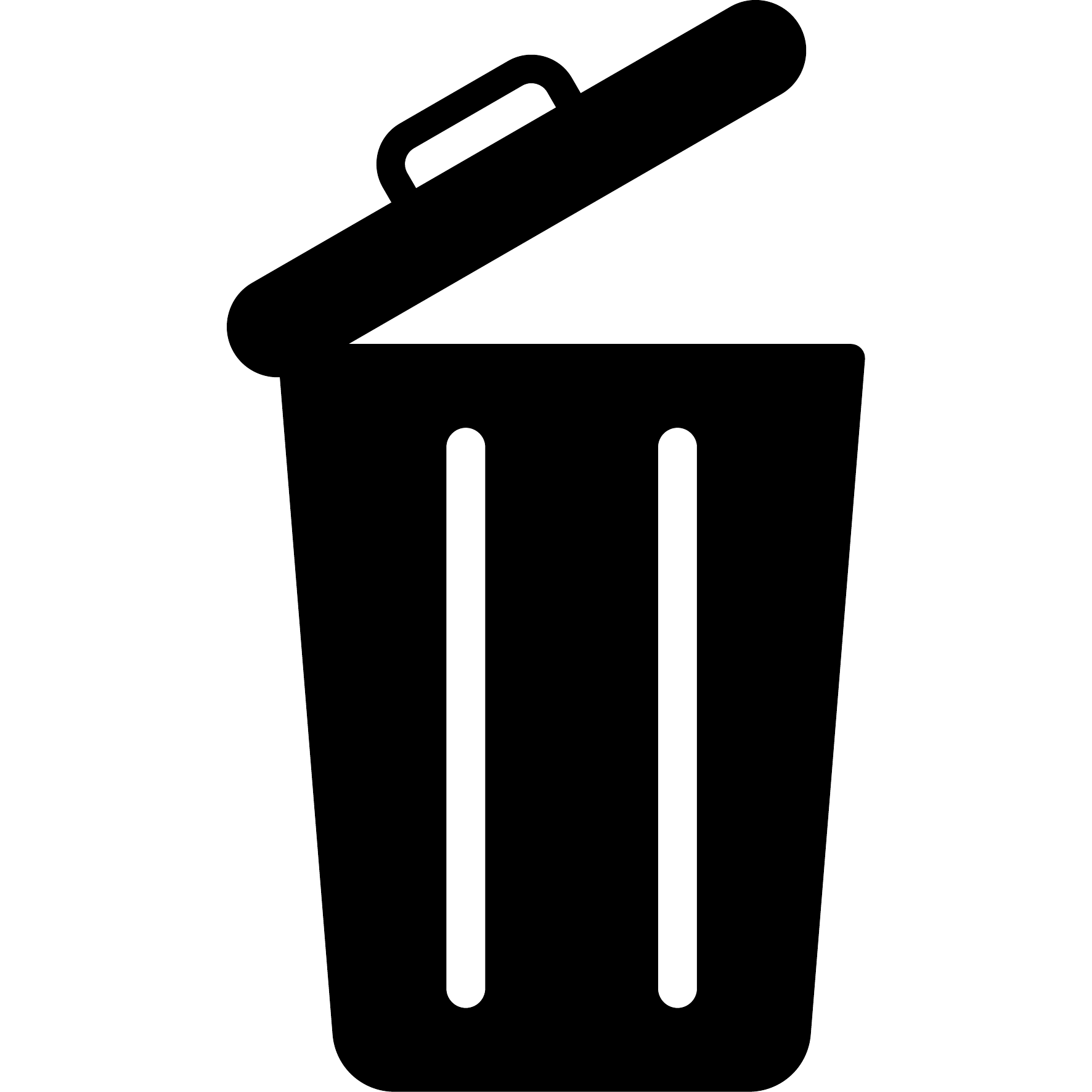}}{b}}
\def\safe{\scalerel*{\includegraphics{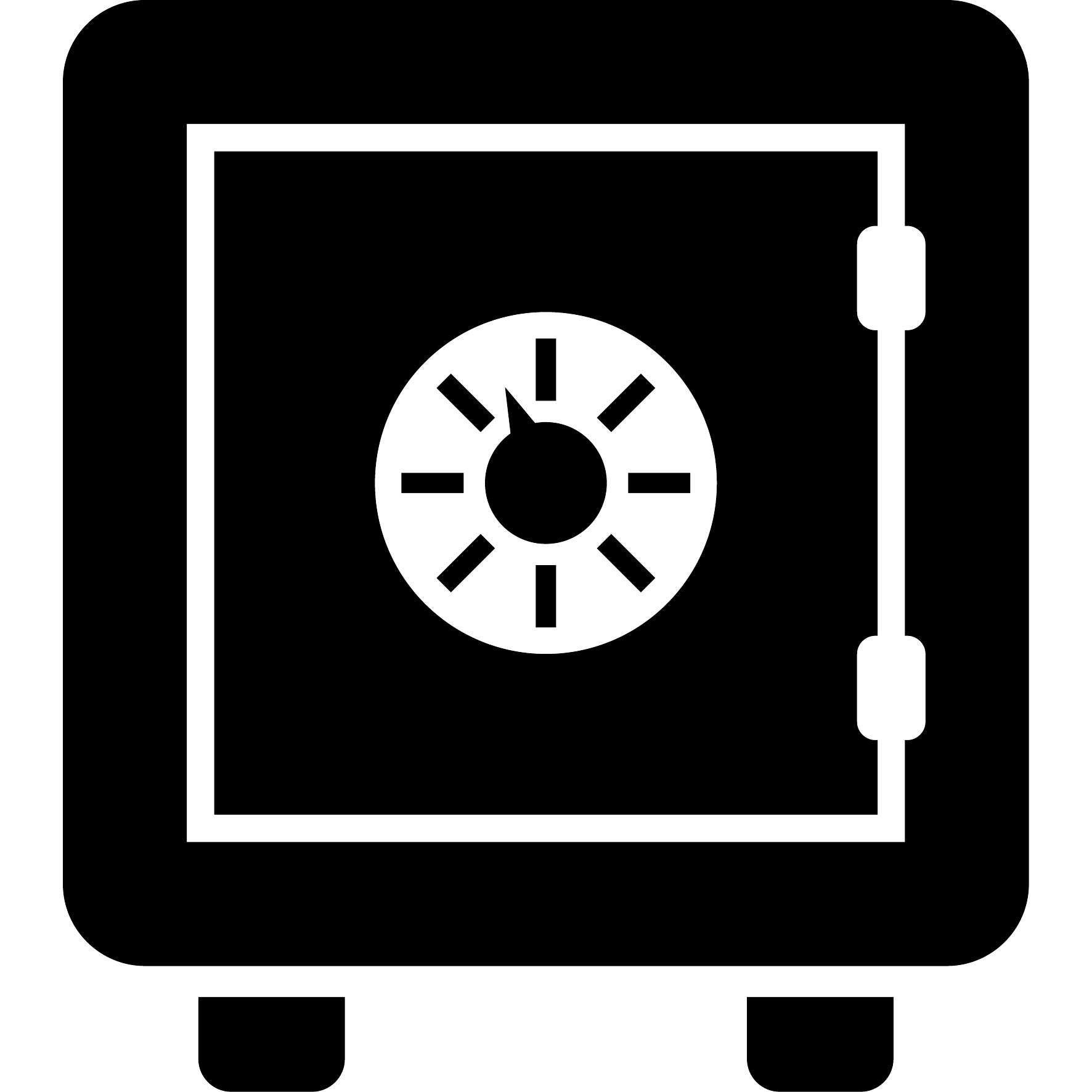}}{b}}
\def\dishwasher{\scalerel*{\includegraphics{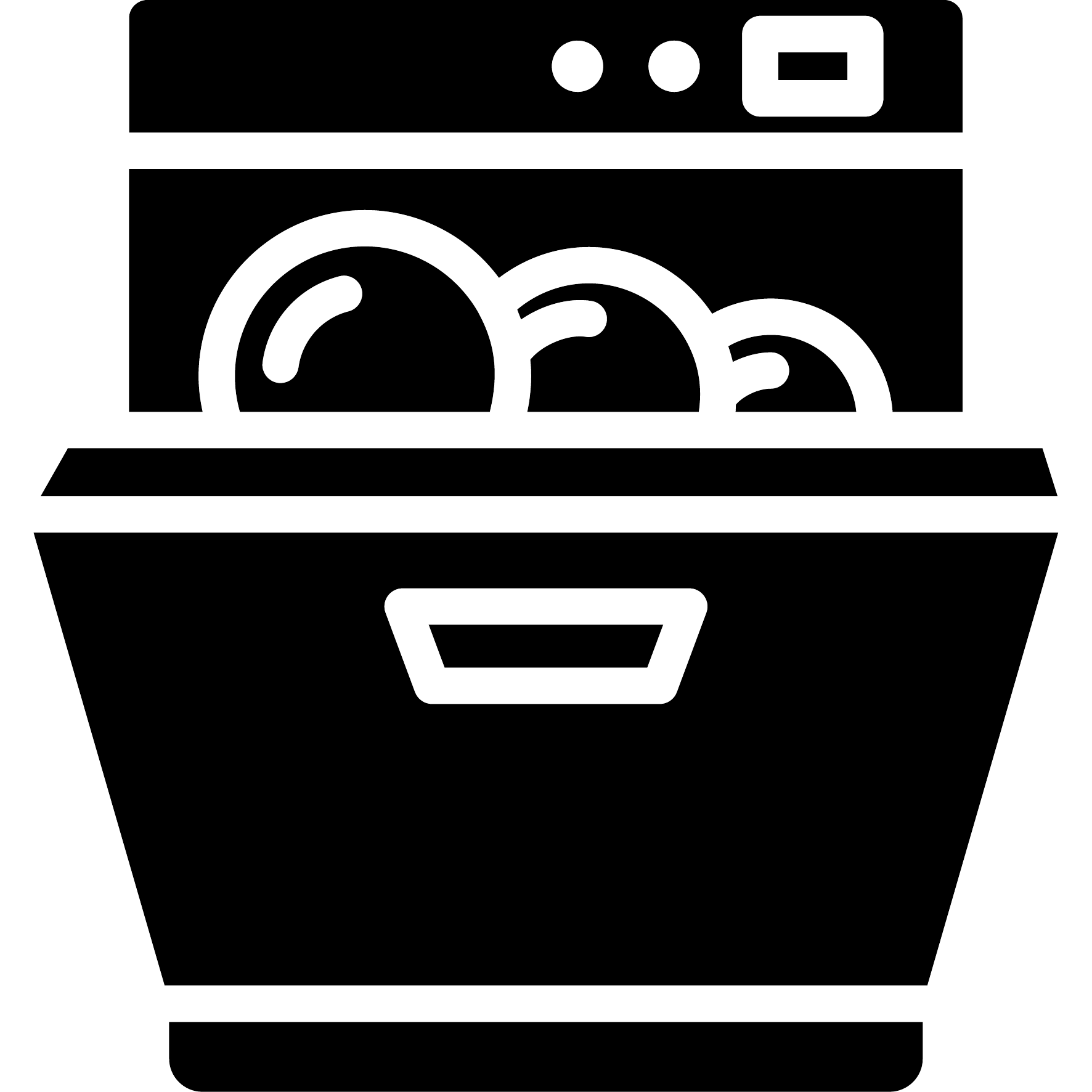}}{b}}
\def\microwave{\scalerel*{\includegraphics{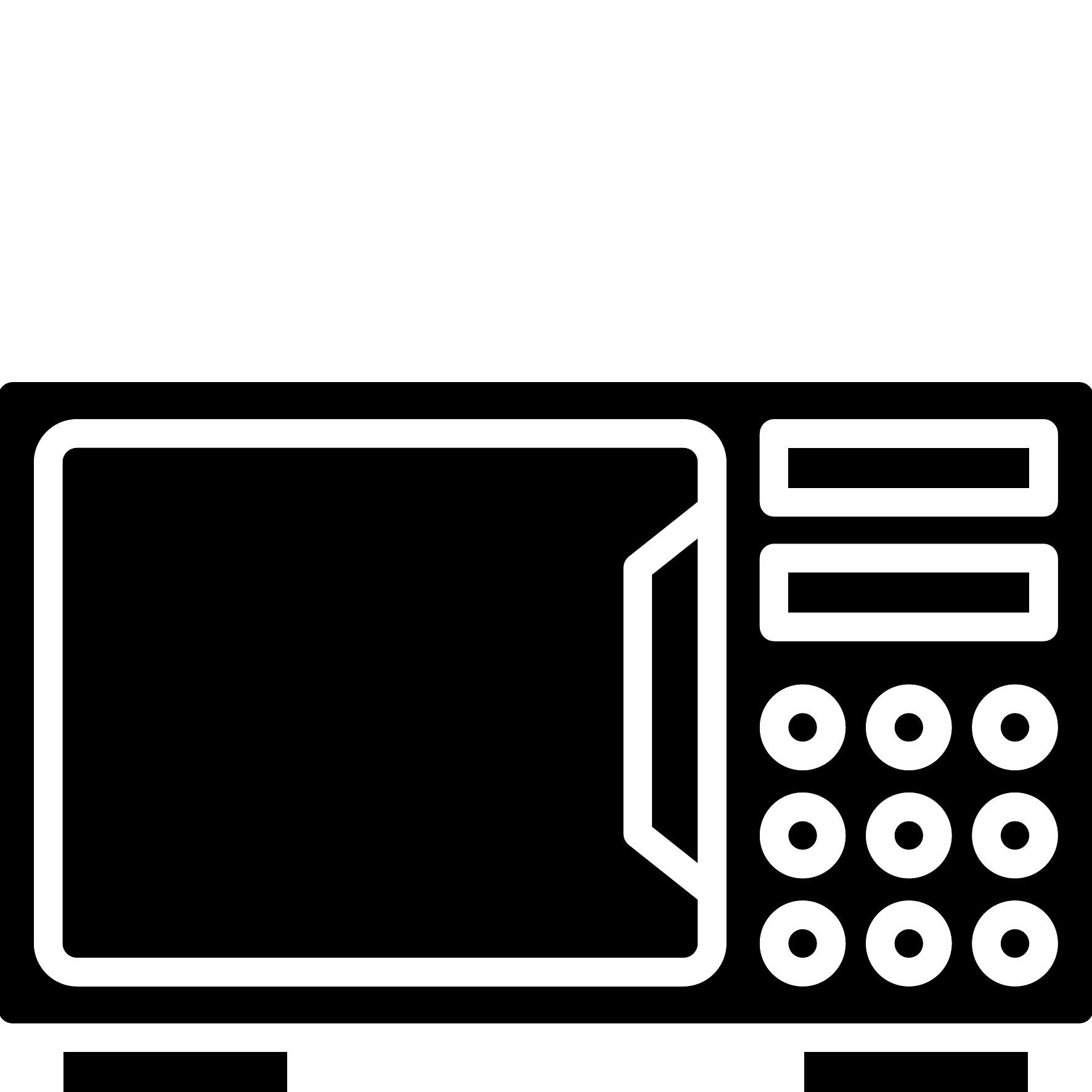}}{b}}
\def\cardbox{\scalerel*{\includegraphics{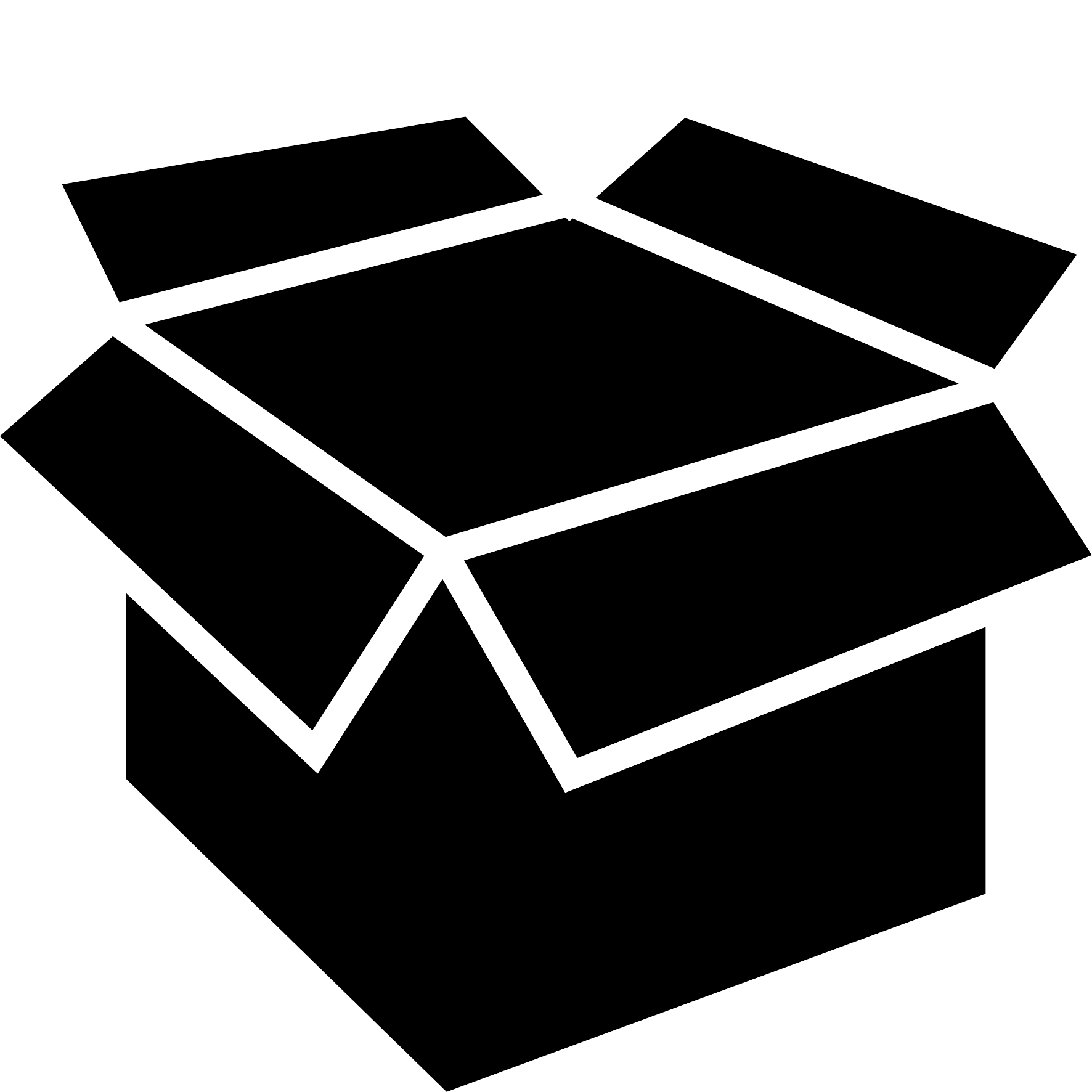}}{b}}
\def\oven{\scalerel*{\includegraphics{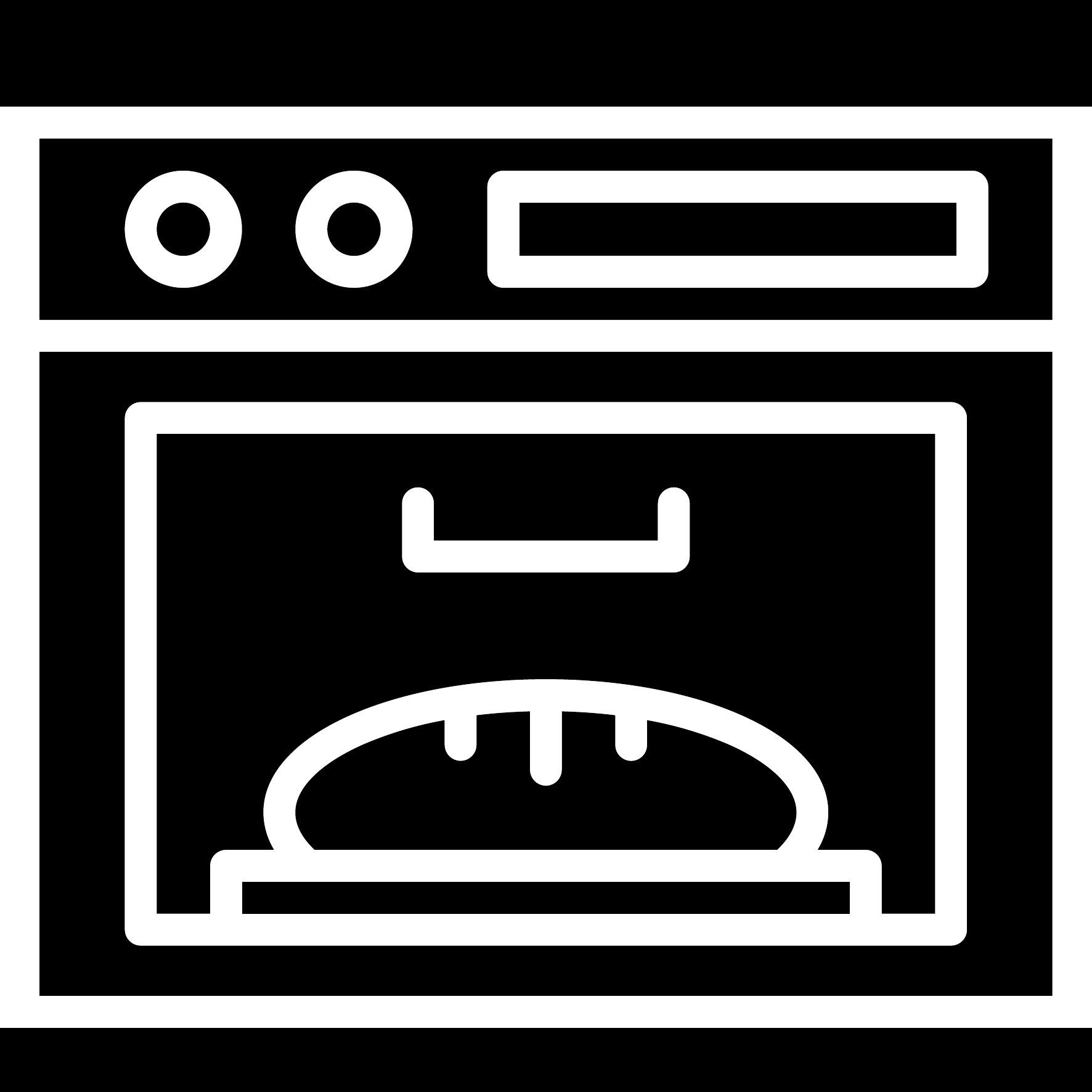}}{b}}
\def\stor{\scalerel*{\includegraphics{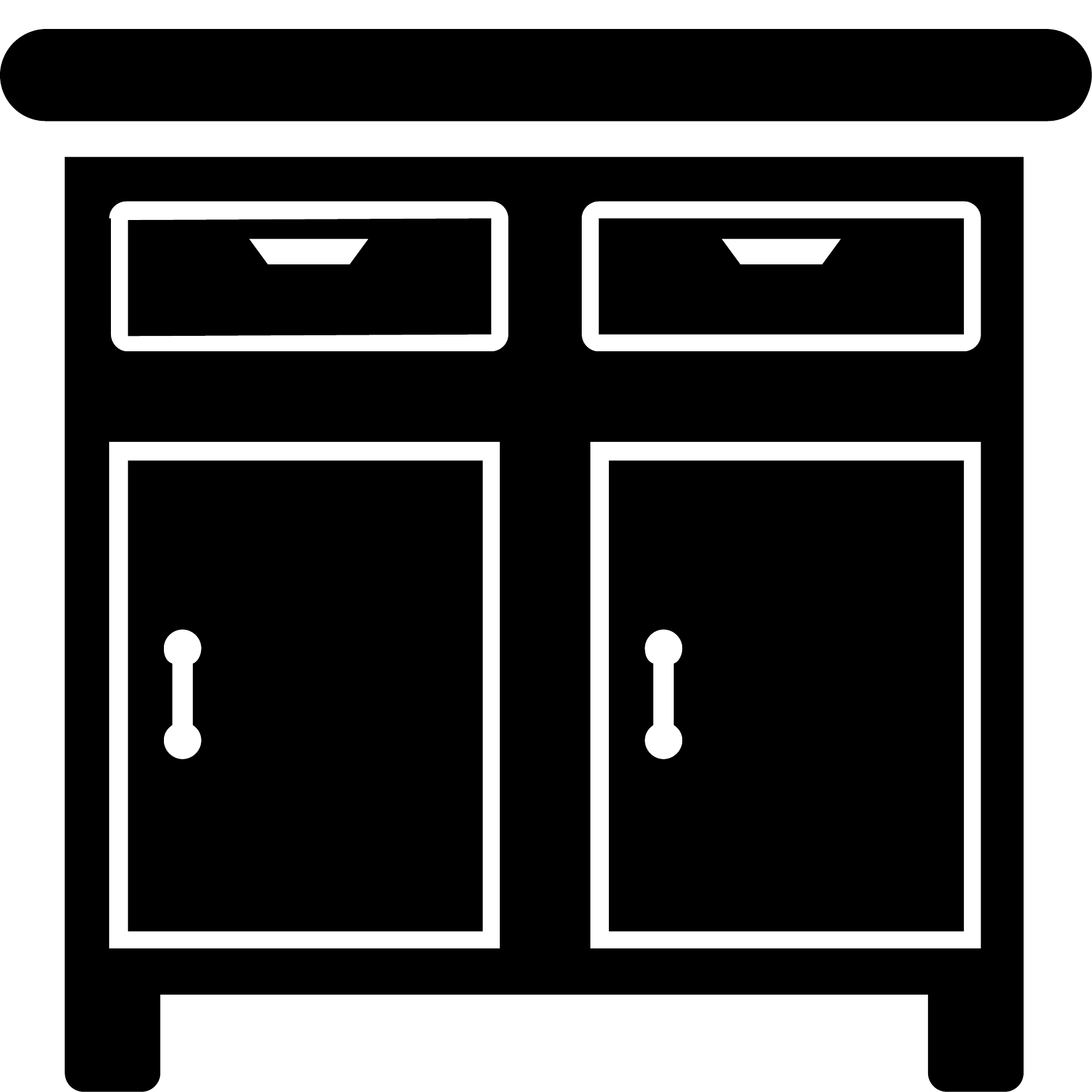}}{b}}

\def\drawer{\scalerel*{\includegraphics{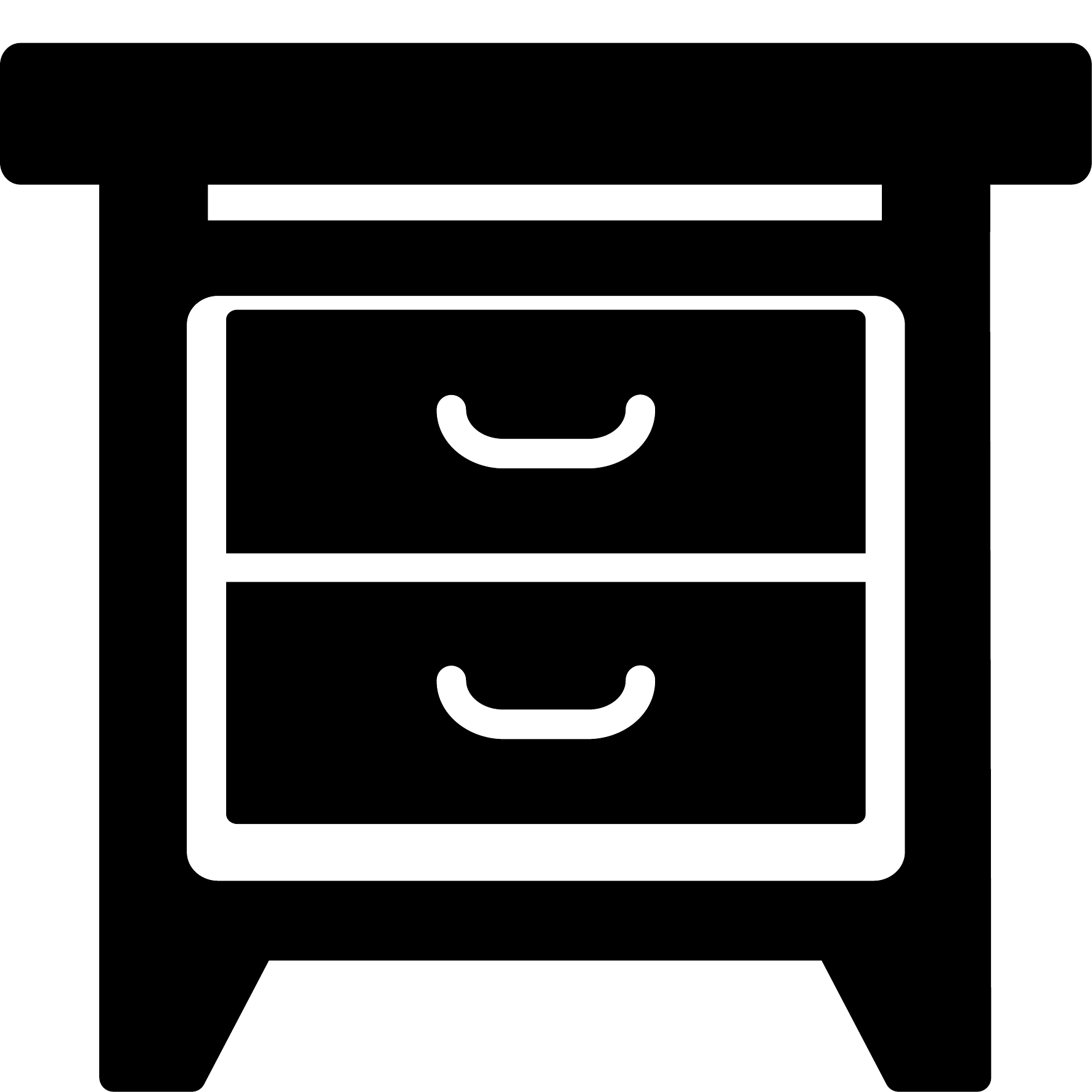}}{b}}
\def\toaster{\scalerel*{\includegraphics{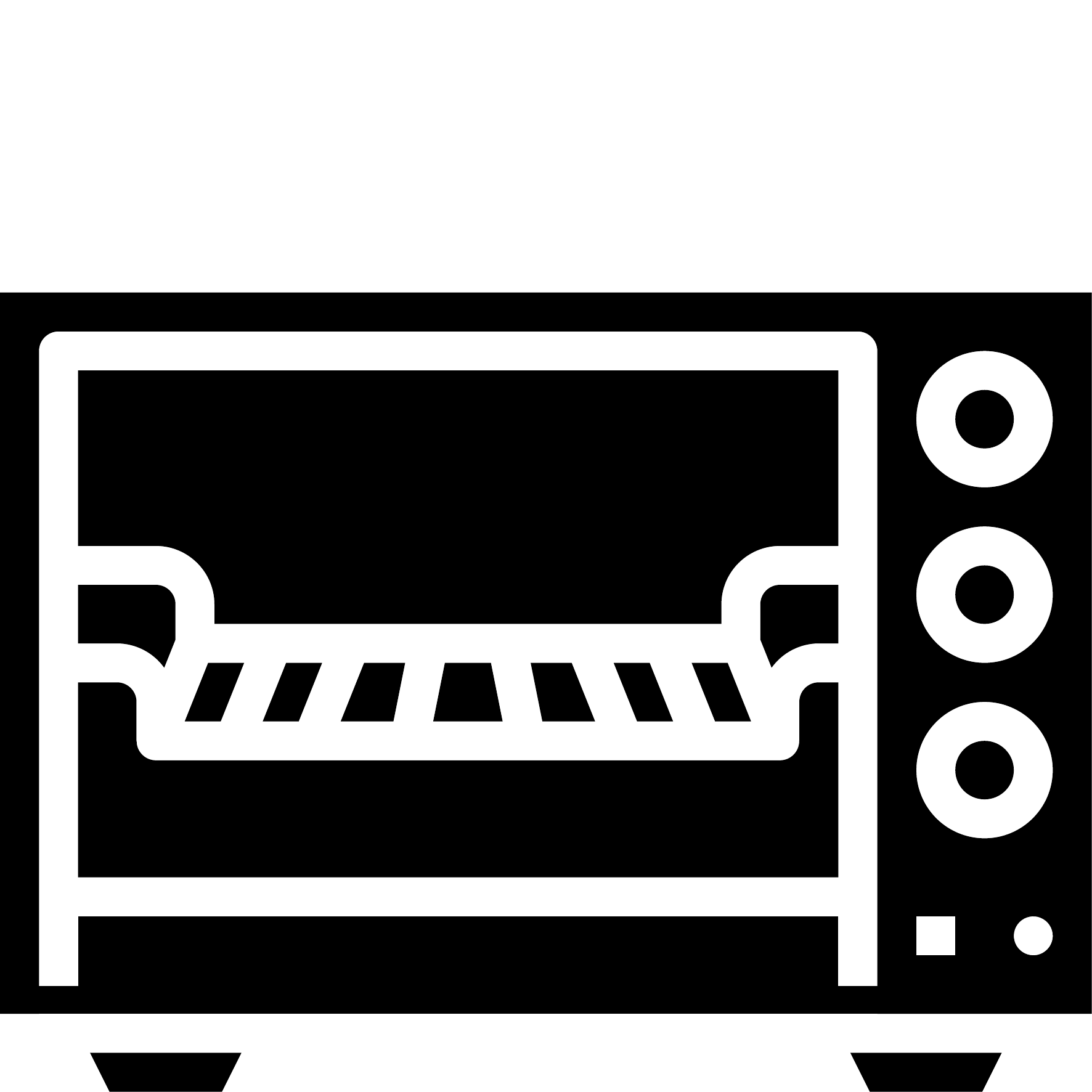}}{b}}
\def\hingeup{\scalerel*{\includegraphics{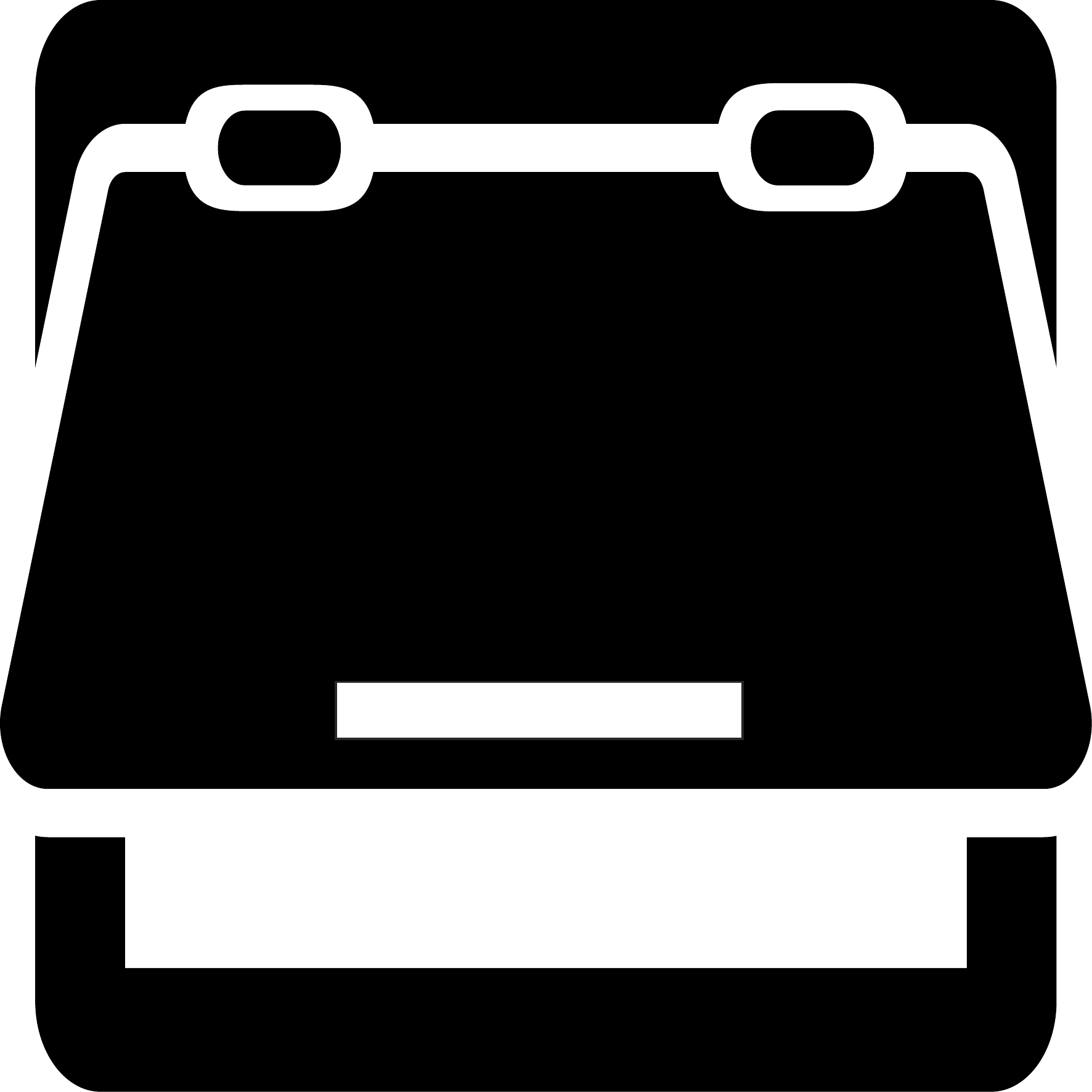}}{b}}
\def\hingedown{\scalerel*{\includegraphics{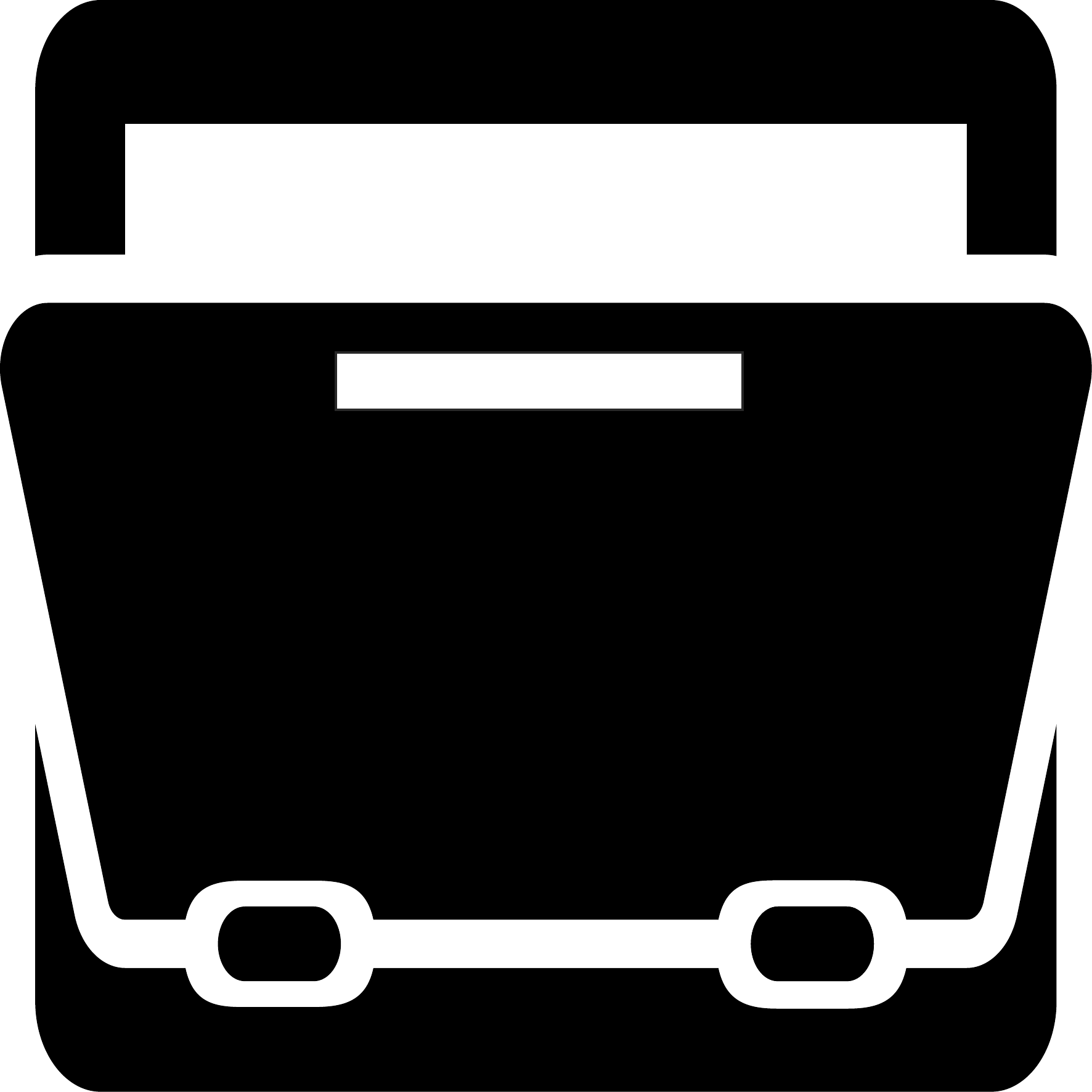}}{b}}
\def\hingeright{\scalerel*{\includegraphics{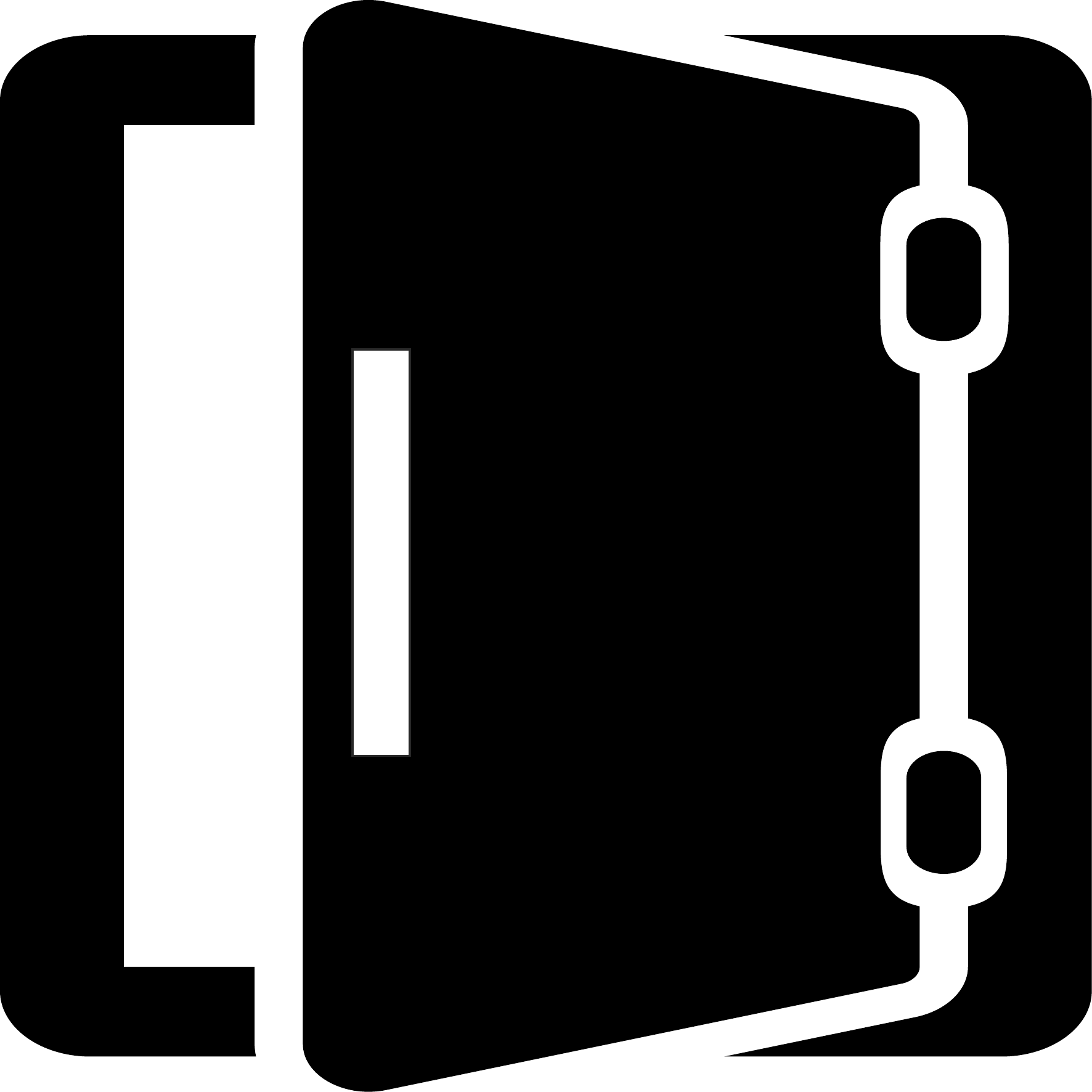}}{b}}
\def\hingeleft{\scalerel*{\includegraphics{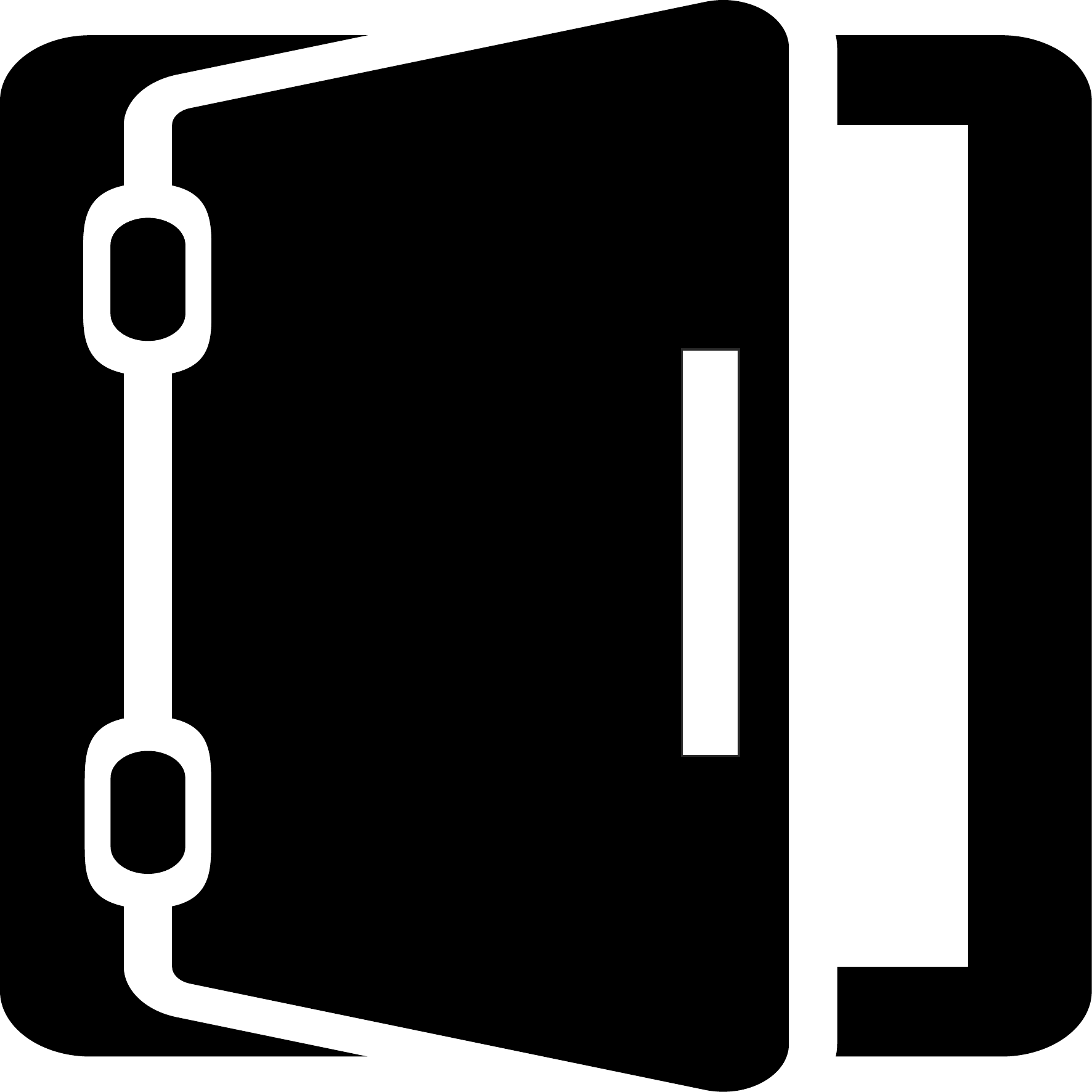}}{b}}

\def\handlecyl{\scalerel*{\includegraphics{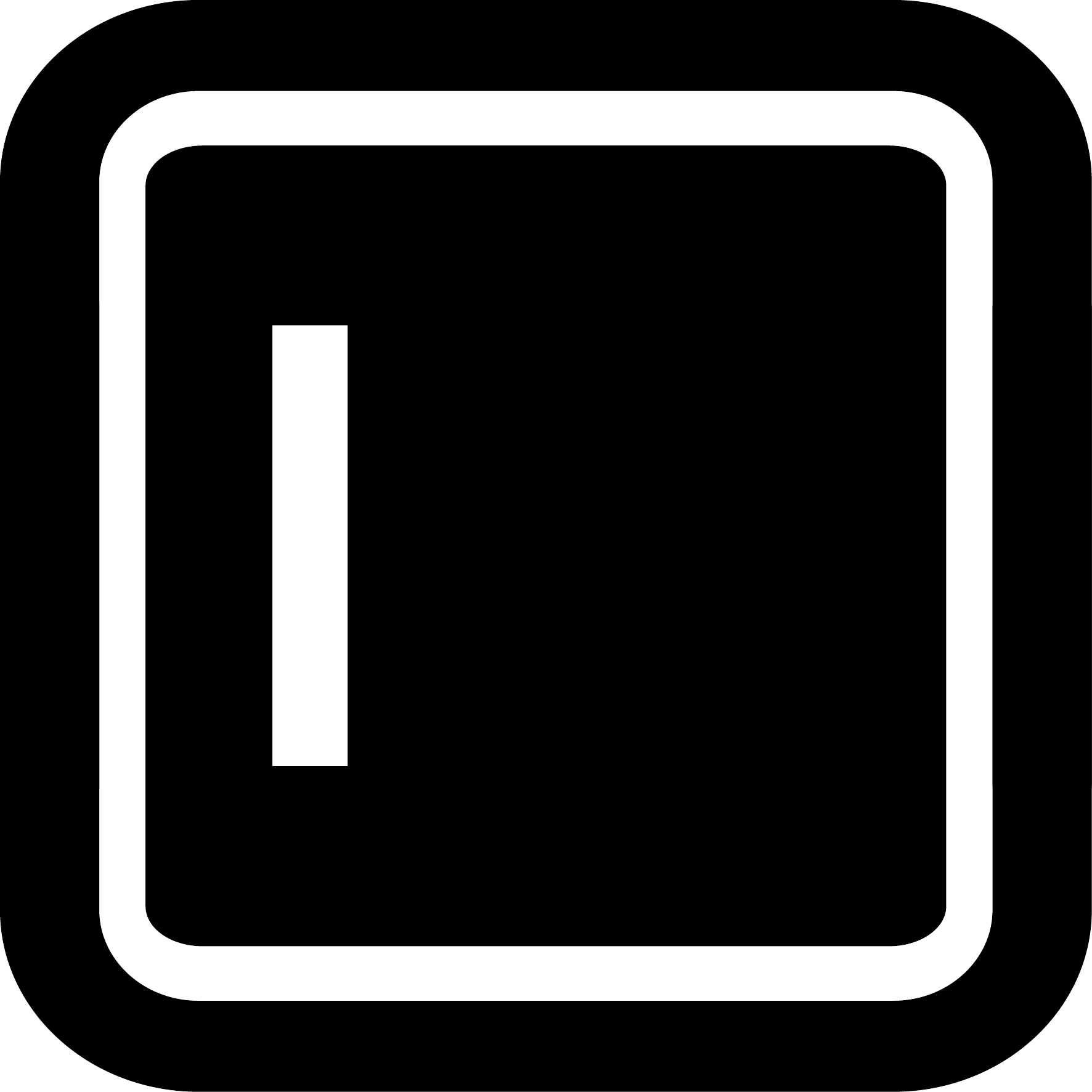}}{b}}
\def\handleL{\scalerel*{\includegraphics{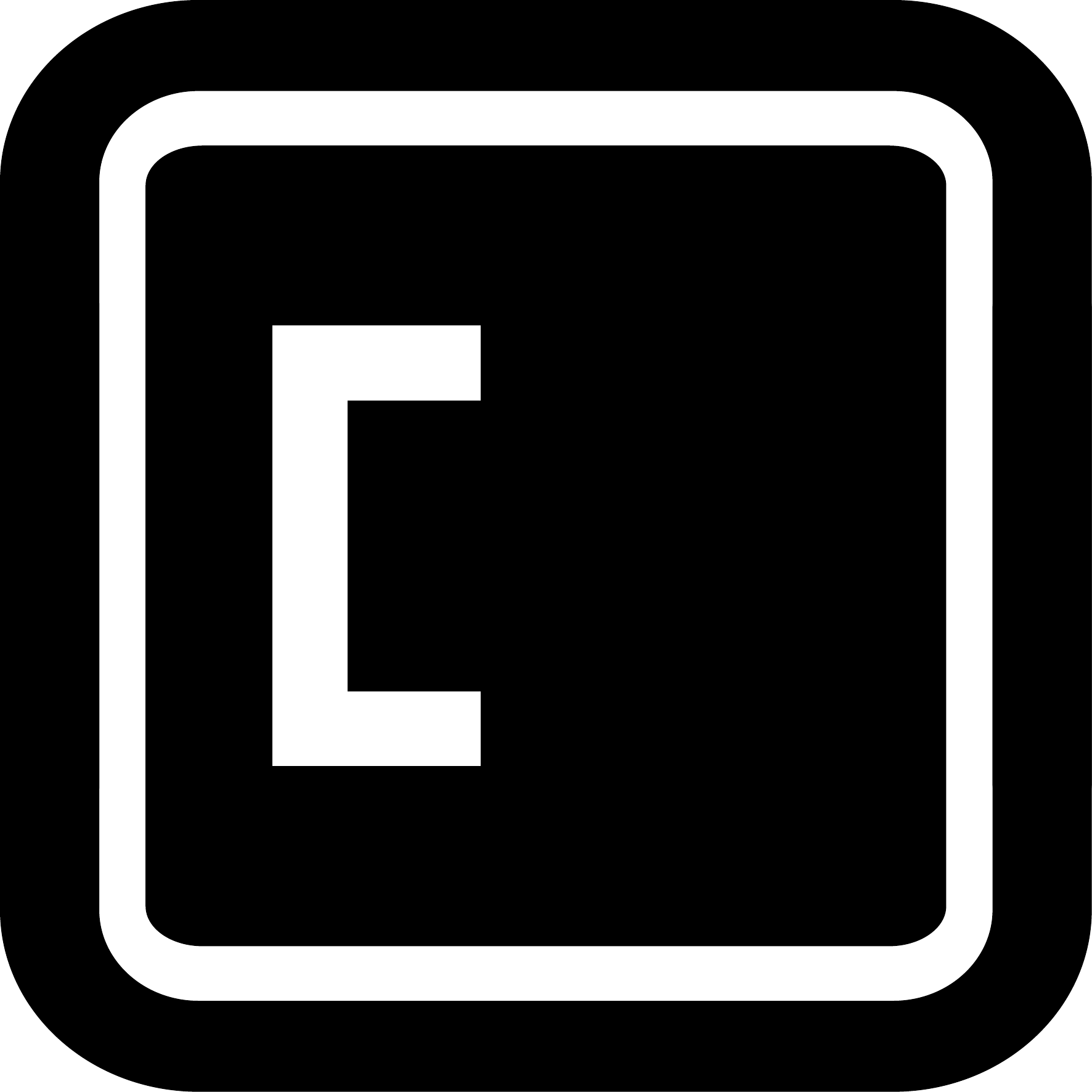}}{b}}
\def\handletex{\scalerel*{\includegraphics{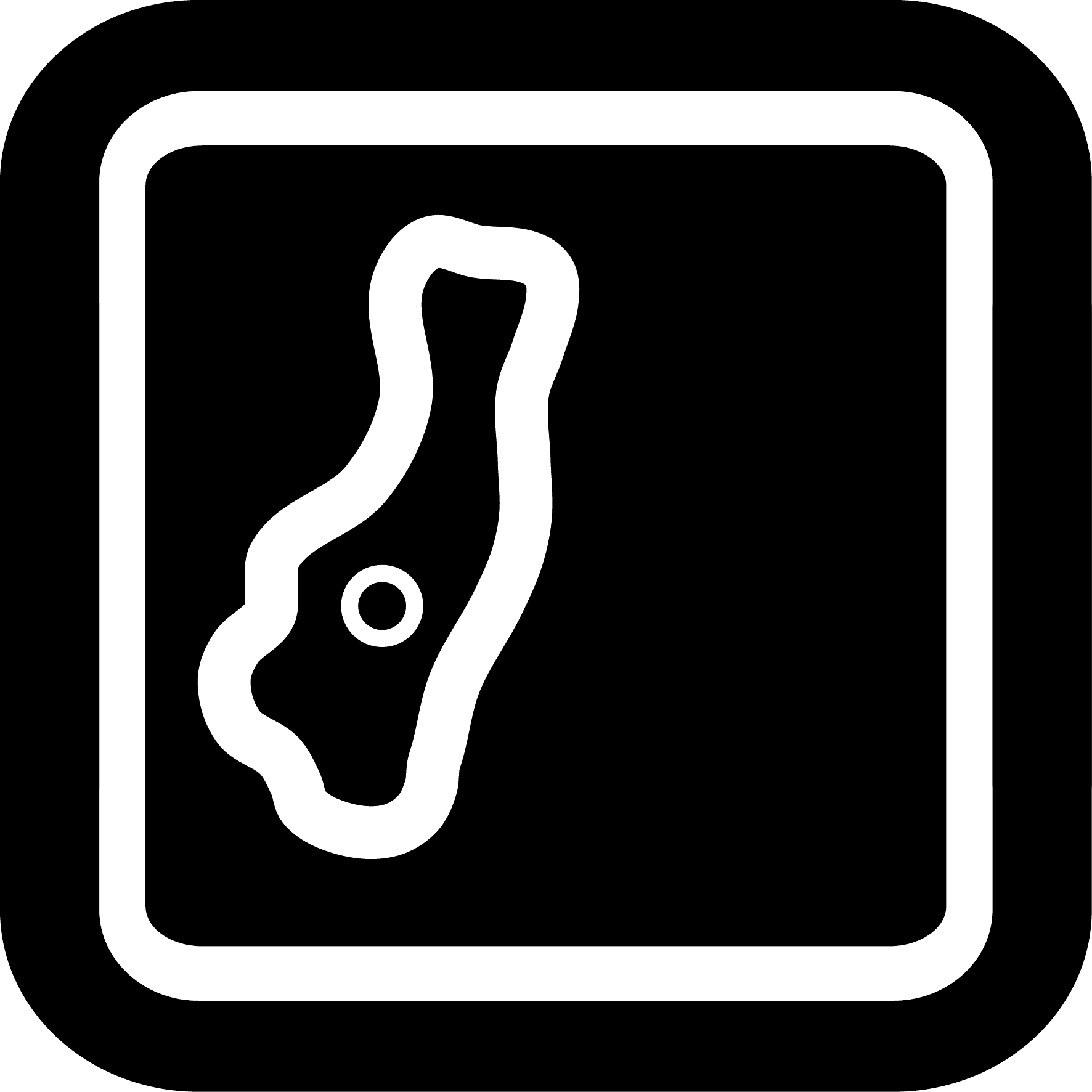}}{b}}
\def\handlesmooth{\scalerel*{\includegraphics{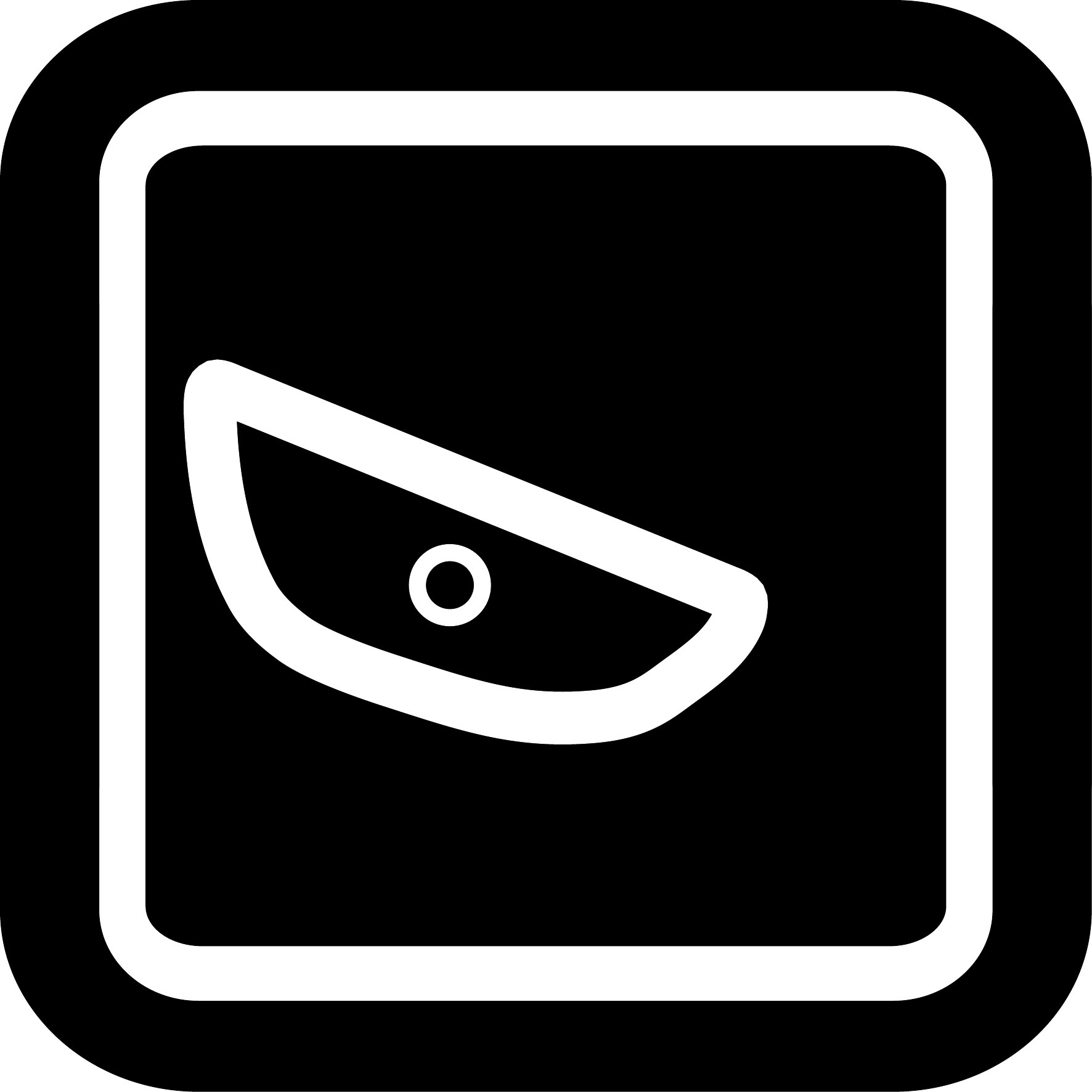}}{b}}

\title{\LARGE \bf
\ourName: \ourNameFull
}

\author{
Carlota Par\'es Morlans$^{*1}$, Claire Chen$^{*1}$, Yijia Weng$^1$, Michelle Yi$^1$, Yuying Huang$^1$, Nick Heppert$^2$,\\Linqi Zhou$^1$, Leonidas Guibas$^1$, and Jeannette Bohg$^1$
\thanks{$^*$Equal contribution.}
\thanks{ $^1$Stanford University, CA, USA. $^2$University of Freiburg, Germany.}
\thanks{\,\,\,Toyota Research Institute provided funds to support this work.}
}
\begin{document}

\maketitle
\thispagestyle{empty}
\pagestyle{empty}

\begin{abstract}

We introduce \ourName{}, a grasp proposal method that generates 6 DoF grasps that enable robots to interact with articulated objects, such as opening and closing cabinets and appliances. \ourName{} consists of two main contributions: the \ourModel{} and the \ourData{}. Given a segmented partial point cloud of a single articulated object, the \ourModel{} predicts the best grasp points on the object with an \modelPointscore{}. Then, it finds corresponding grasp orientations for each of these points, resulting in stable and actionable grasp proposals. We train the \ourModel{} on our new \ourData{}, which contains 78K actionable parallel-jaw grasps on synthetic articulated objects.
In simulation, \ourName{} achieves a 45.0\% grasp success rate, whereas the highest performing baseline achieves a 35.0\% success rate. Additionally, we evaluate \ourName{} on 120 real-world scenes of objects with varied geometries, articulation axes, and joint states, where \ourName{} produces successful grasps on 67.5\% of scenes, while the baseline only produces successful grasps on 33.3\% of scenes. To the best of our knowledge, \ourName{} is the first method for
generating \mbox{6 DoF} grasps on articulated objects directly from partial point clouds without requiring part detection or hand-designed grasp heuristics. The \ourData{} and a pre-trained \ourName{} model are available at our project website: \href{https://stanford-iprl-lab.github.io/ao-grasp/}{https://stanford-iprl-lab.github.io/ao-grasp/}.

\end{abstract}

\section{Introduction}

Human environments are filled with articulated objects, i.e., objects that have movable parts essential to their function, such as storage furniture and appliances. For example, a typical household contains objects like cabinets, dishwashers, and boxes. For robots to autonomously perform tasks in such spaces, they must be able to interact with articulated objects. In this work, we consider the first crucial but challenging step of interacting with any articulated object: determining how a robot can grasp it to enable downstream tasks.

Interacting with articulated objects via grasps allows robots to move the end-effector in any direction without losing contact with an object. In contrast to grasping, non-prehensile manipulation requires the end-effector to move in a limited set of directions in order to maintain contact with the object. Consequently, non-prehensile manipulation constrains the ways in which a robot can interact with articulated objects. Moreover, \cite{Schiavi-2023-AgentAware}, which only considers non-prehensile manipulation, concedes that grasp prediction for articulated objects is important future work because ``not all tasks can be solved through non-prehensile manipulation". Finally, interacting with articulated objects via grasps simplifies actuating the object part after contact is made, because it enables the use of compliant control in a local task frame \cite{Khatib-1987-opspace, Bruyninckx-1996-taskframe, stuede-2019-compliant, meeussen-2010-compliant, prats-2008-compliant}, an approach we describe in \mbox{Section \ref{sec:real-data}}, instead of of planning complex, object-specific trajectories.

Grasping articulated objects presents two unique challenges compared to grasping non-articulated objects. Firstly, grasps not only need to be stable, but also need to be actionable. Grasping any arbitrary point on an articulated object may not be sufficient; the grasp must be on the actuated part of the object to facilitate downstream tasks. For example, to open a microwave, a robot must achieve a stable grasp on the door; grasping the microwave body would be useless. Secondly, a single articulated object can exist in an infinite number of joint configurations, which may have different graspable regions. For example, to open a microwave with a closed door, a robot needs to grasp its handle, while opening the same microwave with an open door also allows the robot to grasp the edge of the door. While there are numerous works on grasping non-articulated, i.e. rigid, objects~\cite{newbury2022review,6672028}, these are not directly applicable to grasping articulated objects because of the two properties described above. Instead, we need to develop grasp prediction methods specific to articulated objects.

\begin{figure}[t]
    \centering
    \includegraphics[width=\linewidth]{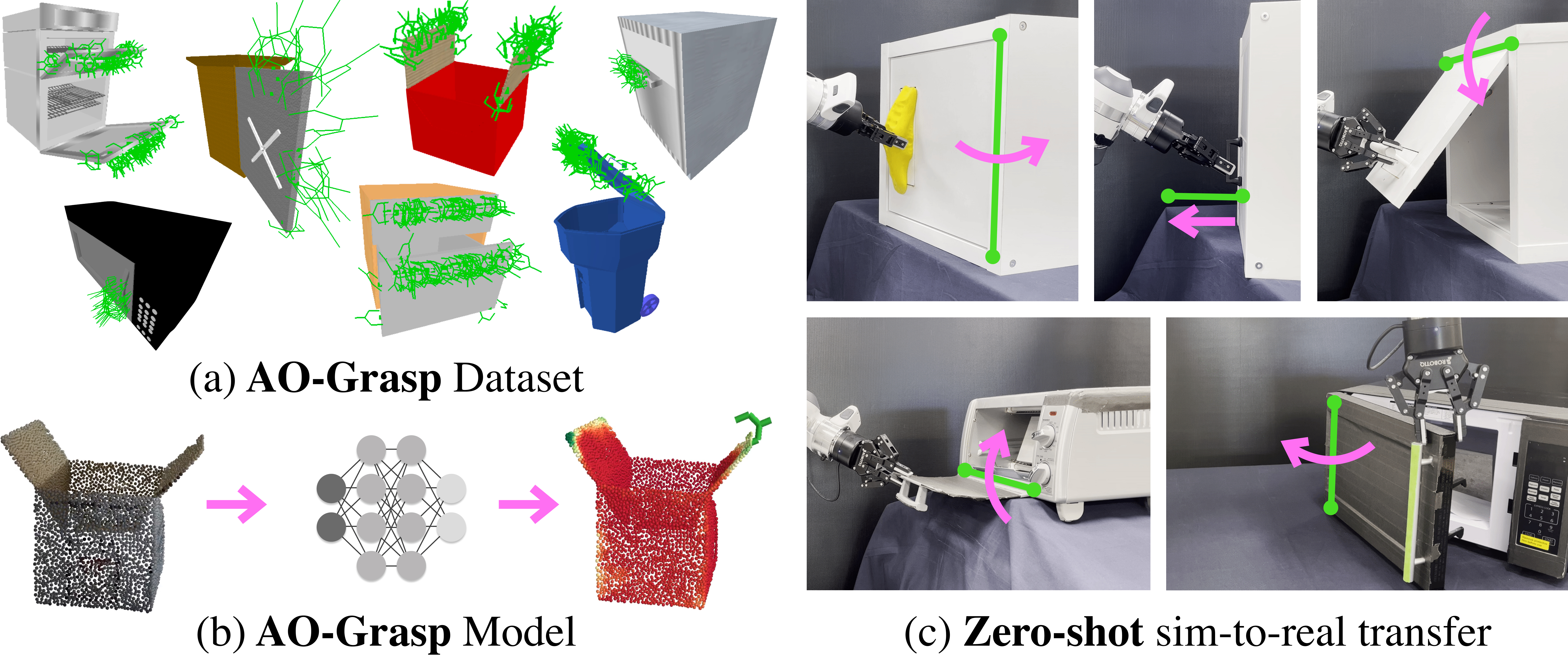}
    \caption{\ourName{} consists of (a) the \ourData{}, which contains 78K actionable grasps on synthetic articulated objects, and (b) the \ourModel{}, which takes a partial point cloud of an articulated object and generates stable and actionable 6 DoF grasps that facilitate downstream manipulation. \ourName{} not only outperforms baselines in simulation, but also achieves zero-shot sim-to-real transfer (c), enabling interactions with real-world objects with different articulation axes and geometries.}
    \vspace{-7mm}
    \label{fig:teaser}
\end{figure}

Guided by this insight, we introduce \ourName{}, which is, to the best of our knowledge, the first method for generating 6 degree-of-freedom (DoF) grasps on articulated objects directly from partial point clouds without requiring part detection or hand-designed grasp heuristics. \ourName{} consists of two main contributions. First, we present the \ourData{} (Fig. \ref{fig:teaser}a) , which contains 78K parallel-jaw 6 DoF grasps on 84 \texttt{Box}, \texttt{Dishwasher}, \texttt{Microwave}, \texttt{Safe}, \texttt{TrashCan}, \texttt{Oven}, and \texttt{StorageFurniture} instances from the PartNet-Mobility dataset~\cite{Xiang_2020_SAPIEN,Mo_2019_CVPR}. The \ourData{} only contains grasps on actionable parts of articulated objects, as well as grasps on objects in diverse joint configurations. Next, we introduce the \ourModel{} (Fig. \ref{fig:teaser}b), which generates stable and actionable 6 DoF parallel-jaw grasps for articulated objects given a partial point cloud of an object. The \ourModel{} predicts where on objects a robot should grasp with an \modelPointscore{}, trained on the \ourData{}. Unlike other works \cite{Mo-2021-W2A, Schiavi-2023-AgentAware, Wu_2022-VATMART}, the \modelPointscore{} does not require additional semantic part segmentation to predict grasp points on actionable parts of articulated objects.
Finally, we generate grasp orientations for the top predicted grasp points.

\begin{figure*}[h]
    \centering
    \includegraphics[width=\linewidth]{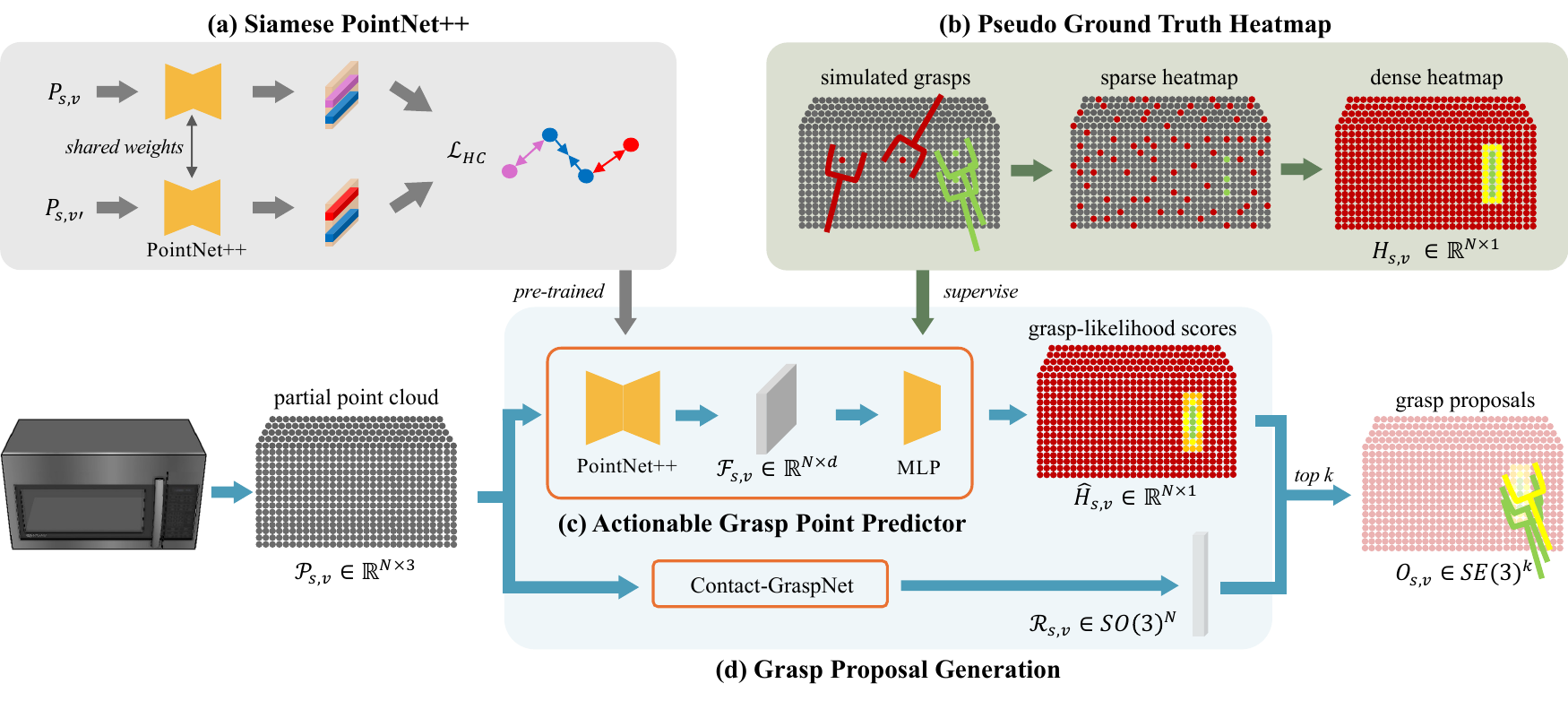}
    \caption{An overview of the \ourModel{}. (a) Siamese PointNet++: We find positive and negative correspondences between partial point clouds of an object in joint state $s$ captured from two different object views $v$ and $v'$ to train the network with a hardest contrastive loss $L_{HC}$ (\ref{eq:hc}). (b) Pseudo ground truth heatmap $H_{s,v}$ (\ref{eq:heatmap}): We supervise training the \modelPointscore{} on dense heatmaps computed from grasps in the \ourData{}. (c) Actionable Grasp Point Predictor: Given a partial point cloud of an articulated object $P_{s,v}$, the AO-Grasp Point Predictor outputs a heatmap of grasp-likelihood scores $\hat{H}_{s,v}$. It first processes the partial point cloud $P_{s,v}$ with the pre-trained PointNet++ module, resulting in point features $F_{s,v}$. Then, these features are passed through an MLP which returns the grasp-likelihood scores $\hat{H}_{s,v}$. (d) Grasp Proposal Generation: We leverage Contact-GraspNet \cite{Sundermeyer-2021-CGN} to get per-point grasp orientations. We take the top $k$ scores from $\hat{H}_{s,v}$ to get a set of stable and actionable 6 DoF grasp poses $O_{s,v} \in SE(3)^k$.}
    \label{fig:pipeline}
    \vspace{-5mm}
\end{figure*}

We show, in simulation and through zero-shot transfer to the real-world (Fig. \ref{fig:teaser}c), that \ourName{} generates actionable grasps on articulated objects with diverse handle geometries and articulation axes, even on categories not seen in training. \ourName{} also generates grasps for objects in both closed and open joint configurations and viewed from varied camera angles. In simulation, we show that \ourName{} achieves an average $45.0\%$ success rate across unseen instances from both train and test categories, while a rigid object grasping baseline CGN \cite{Sundermeyer-2021-CGN} and articulated object interaction baselines VAT-Mart \cite{Wu_2022-VATMART} and Where2Act \cite{Mo-2021-W2A} achieve average success rates of $35.0\%$, $5.21\%$, and $3.9\%$ respectively. In the real-world, we compare \ourName{} and CGN on a custom-built re-configurable cabinet with 4 handles and 4 articulation axes, 2 microwaves, a toaster oven, prismatic-jointed drawer, and cardboard box. On 120 real-world scenes, \ourName{} produces successful grasps on 67.5\% of scenes, while CGN only produces successful grasps on 33.3\% of scenes.%

\section{Related work}
This work focuses on interacting with articulated objects via grasps. As such, we situate this work amongst both the literature on interacting with articulated objects and the body of work on grasping rigid objects. We also differentiate the \ourData{} from existing grasp datasets.

{\noindent \textbf{Interacting with articulated objects}:}
While there are many existing works on interacting with articulated objects, none specifically addresses the task of generating stable grasps for objects in diverse configurations without requiring part detection or hand-designed grasp heuristics.%

Some works consider interacting with articulated objects via grasps \cite{liu_v-mao_2021, Geng-2023-gapartnet, Zhang-2023-flowbot, Klingbeil-2010-LearningTo, Arduengo-2019-AVF, Ruhr-2012-AGF, stuede-2019-compliant, xiong2024adaptive}, but use various strategies to simplify grasp generation. Unlike \ourName{}, V-MAO \cite{liu_v-mao_2021} only predicts contact points, as opposed to 6 DoF poses, and requires part segmentation of the object, whereas \ourName{} does not. V-MAO also does not show real-world results. Other works use hand-designed grasp heuristics \cite{Geng-2023-gapartnet, Zhang-2023-flowbot, Klingbeil-2010-LearningTo, Arduengo-2019-AVF, Ruhr-2012-AGF, stuede-2019-compliant, xiong2024adaptive} that would not scale well to a larger variety of objects, motivating a more general grasp generation method like \ourName{}.

Another category of work focuses on learning interaction policies that do not require grasps \cite{Mo-2021-W2A, Wu_2022-VATMART, Geng-2023-gapartnet, Geng-2023-partmanip, bao_dexart_2023, Schiavi-2023-AgentAware, Curtis-2023-pomdp, Xu-2022-UMPNet, eisner2023flowbot3d, ning2023where2explore}. Where2Act (W2A) \cite{Mo-2021-W2A} and VAT-Mart \cite{Wu_2022-VATMART} are most similar to \ourName, as they also predict per-point interaction poses given partial point clouds. Unlike \ourName, however, W2A and VAT-Mart do not specifically check that the robot achieves stable grasps on objects, resulting in interactions that are predominantly non-prehensile. In our evaluation, we confirm that these methods rarely produce successful grasps, which limits how a robot can interact with an articulated object. Some of these works \cite{Xu-2022-UMPNet, Zhang-2023-flowbot} even further simplify making contact with the object by using a suction cup end-effector, which constrains the types of geometries the robot can interact with to smooth and flat surfaces. Furthermore, although these works all showcase interactions on an impressive variety of articulated objects in simulation, most only show limited or no real-world results. Of the works that do include real-world results \cite{Xu-2022-UMPNet, Geng-2023-gapartnet, Geng-2023-partmanip, Wu_2022-VATMART, Schiavi-2023-AgentAware, ning2023where2explore, eisner2023flowbot3d}, only one \cite{eisner2023flowbot3d}, which uses a suction cup, conducts quantitative real-world evaluations. 
In contrast, we show that \ourName{} can be used in the real-world by quantitatively evaluating \mbox{\ourName{}} on 16 variations of a reconfigurable cabinet and 5 additional common household articulated objects.

{\noindent \textbf{Datasets}:}
A line of work \cite{goldfeder2009columbia,kappler2015leveraging} has been devoted to building large-scale grasp datasets, as they are essential for data-driven grasping approaches. Most related to us are datasets for parallel-jaw grippers \cite{mahler2017dex,mousavian20196,morrison2020egad,fang2020graspnet,eppner2022billion,eppner2021acronym}, which label grasps either analytically \cite{mahler2017dex,morrison2020egad,fang2020graspnet} or by running simulation trials \cite{mousavian20196,eppner2021acronym,eppner2022billion}. However, these datasets only target rigid objects and focus on grasp stability, not actionability. Both W2A \cite{Mo-2021-W2A} and VAT-Mart \cite{Wu_2022-VATMART} release data generation methods for 6 DoF interactions with articulated objects, but these interactions do not guarantee stable grasps, limiting how a robot can interact with objects. To the best of our knowledge, we are the first to build a grasp dataset for articulated objects that contains both stable and actionable grasps.

{\noindent \textbf{Grasping non-articulated objects}:}
Despite the unique challenges that come with grasping articulated objects, an important part of the grasping problem remains the same as in grasping rigid objects: understanding the local geometries of an object that make those areas suitable for grasping. There is a deep body of work studying this grasping problem~\cite{6672028,newbury2022review}. Contact-GraspNet (CGN) \cite{Sundermeyer-2021-CGN} and AnyGrasp \cite{fang2023anygrasprobustefficientgrasp} are two recent works that use large datasets to learn grasp prediction models. We find that once we have predicated good locations at which to grasp articulated objects, we can leverage works like these to predict grasp orientations that match local geometry. \ourName{} uses CGN \cite{Sundermeyer-2021-CGN}, as it is a representative rigid object grasp generation method.

\section{\ourData{}}\label{sec:data-gen}
We introduce the \ourData{}, a dataset of simulated, actionable grasps on articulated objects. It contains 78K 6 DoF grasps for 84 instances from 7 common household furniture/appliance categories (\texttt{Box}, \texttt{Dishwasher}, \texttt{Microwave}, \texttt{Safe}, \texttt{TrashCan}, \texttt{Oven}, \texttt{StorageFurniture}) from the PartNet-Mobility dataset \cite{Mo_2019_CVPR, Xiang_2020_SAPIEN}. Table \ref{tab:raw-dataset} summarizes the per-category statistics of our dataset. Across its 7 categories, the \ourData{} exhibits considerable variation in geometries, articulation axes, joint configurations, and number of actionable parts, as shown in Fig. \ref{fig:teaser}a.

\begin{table}[t]
\footnotesize
\centering
\vspace{2.5mm}
\setlength{\tabcolsep}{3pt} %
\begin{tabular}{ccccccccc} 
\toprule
\textbf{Category} & \textbf{All} & \huge{\cardbox} & \huge{\dishwasher} & \huge{\microwave} & \huge{\safe} & \huge{\trash} & \huge{\oven} & \huge{\stor} \\ 
\midrule
\textit{\# Instances} & \textit{84} & \textit{9} & \textit{17} & \textit{11} & \textit{11} & \textit{13} & \textit{6} & \textit{17} \\
\midrule
\makecell{Closed state\\grasps} & 15632 & 516 & 1396 & 1546 & 372 & 2493 & 2715 & 6594 \\
\makecell{Open state\\grasps} & 62641 & 8091 & 8020 & 8022 & 6152 & 11669 & 5400 & 15287\\
\midrule
Total grasps & 78273 & 8607 & 9416 & 9568 & 6524 & 14162 & 8115 & 21881 \\
\bottomrule
\end{tabular}

\caption{The number of instances and grasps for \texttt{Box}, \texttt{Dishwasher}, \texttt{Microwave}, \texttt{Safe}, \texttt{TrashCan}, \texttt{Oven}, and \texttt{StorageFurniture} categories in the \ourData{}. In total, the \ourData{} spans 7 categories and contains 78K grasps for 84 instances, each in 1 closed state and 9 randomly-sampled states.
}
\vspace{-5mm}
\label{tab:raw-dataset}
\end{table}

\subsection{Grasp parametrization and labeling criteria}\label{sec:grasp-ep}

The \ourData{} uses a two-fingered Robotiq 2F-85 gripper. A grasp is parameterized by a 6 DoF pose $g = (t, R) \in SE(3)$, with a position $t$ and rotation $R$. 

In contrast to rigid object grasping, where the stability of grasps is usually verified by shaking objects or applying disturbance forces \cite{eppner2021acronym,mousavian20196,eppner2022billion}, we require semantically meaningful interactions with articulated objects, such as opening the door of a microwave. Consequently, we design our grasp evaluation procedure to test not only for a grasp's stability, but also its actionability.

We label each grasp by executing a grasp episode with a floating gripper in PyBullet \cite{coumans2016pybullet}. We first spawn the fully-open gripper at $g$, close the gripper to complete the grasp if no collision is detected, then start action execution by moving the gripper in the optimal direction to actuate the object part, which we obtain via the object's ground-truth joint type and axis. After a fixed number of steps, we terminate the action and label the grasp as successful if 1) the gripper is still in contact with the object, indicating stability and 2) the grasped part has moved a certain distance, indicating actionability.

\subsection{Grasp sampling}

For each object instance, we randomly sample grasp poses and label each one using the procedure described in the previous section. For each instance, we generate grasps for the object in its canonically-closed state and 9 randomly-sampled open states, and capture each state from 20 randomly-sampled camera viewpoints. In total, we obtain 78,273 positively-labeled grasps on 84 instances.

\section{\ourModel{}}\label{sec:pipeline}

The \ourModel{} takes a partial point cloud of an articulated object as input and outputs a set of 6 DoF grasp poses (see Fig. \ref{fig:pipeline}). First, it determines where on the object a robot should grasp with the novel \modelPointscore{}, trained on the \ourData{} to predict per-point \scoreName{} scores. We design the \modelPointscore{} loss and training strategies to facilitate generalization to new viewpoints, joint states, and objects. Then, to generate grasp orientations for grasp points predicted by the \modelPointscore{}, the \ourModel{} leverages rotations generated by Contact-GraspNet \cite{Sundermeyer-2021-CGN}, a state-of-the-art rigid object grasping method. Finally, from the per-point \scoreName{} scores and grasp orientations, we compose the final set of grasp proposals by selecting the points with the highest \scoreName{} scores.

\subsection{\modelPointscore{}}
\label{sec:training_strategies}
For each point in the input point cloud, the \modelPointscore{} outputs a grasp-likelihood score that signifies how likely that point will afford a stable and actionable grasp. It consists of a PointNet++ (PN++) \cite{Qi-2017-PNpp} backbone, which extracts per-point features, and an MLP output head that learns to predict the grasp-likelihood scores of points given their features. We use the following 2 strategies to achieve generalization across viewpoints, joint states, and object instances and categories:

{\noindent\textit{1) Learning viewpoint-independent point correspondences}}:

To achieve generalization across different camera viewpoints, the \modelPointscore{} must understand that grasp-likelihood scores are viewpoint-independent. To facilitate this understanding, we pre-train PN++ using a Siamese Network in a self-supervised manner using the hardest-contrastive loss proposed in \cite{choy2019fully}. We finetune this pre-trained PN++ backbone when training the grasp-likelihood output head.

Inspired by \cite{zhou2020siamesepointnet}, we use a Siamese Network architecture to learn viewpoint-independent point-wise features (see Fig. \ref{fig:pipeline}a). One training pass of the Siamese Network takes, as input, two partial view point clouds, $\mathcal{P}_{s, v}$ and $\mathcal{P}_{s,v'}$, of a single instance in the same joint state $s$, but captured from different camera viewpoints $v$ and $v'$. These are passed through PN++ to compute point-wise features $F_{s,v}$ and $F_{s,v'}$. Next, we uniformly sample a set $Z$ of pairs of matched points $(i, j)$ where $i \in {P}_{s, v}$ and $j \in {P}_{s, v'}$, with corresponding features $f^i_{s,v} \in F_{s,v}$ and $f^{j}_{s,v'} \in F_{s,v'}$. Then, we compute the following hardest contrastive loss across all point pairs,

{\small
\begin{dmath}\label{eq:hc}
 \mathcal{L}_{HC}=\sum_{(i, \,j) \in\mathcal{Z}}\left [\frac{\left [\left \| f^{i}_{s,v}-f^{j}_{s,v'} \right \|_2 - m_p \right ]^2_+}{\left | \mathcal{Z} \right |} + \frac{\left [ m_n-\min_{k\in \mathcal{N}_{s,v'}}\left \| f^{i}_{s,v}-f^{k}_{s,v'} \right \|_2 \right ]^2_+}{2\left |\mathcal{N}_{s,v'}\right |} + \frac{\left [m_n-\min_{k\in \mathcal{N}_{s,v}}\left \| f^{j}_{s,v'}-f^{k}_{s,v} \right \|_2\right ]^2_+}{2\left |\mathcal{N}_{s,v}\right |}\right ]
\end{dmath},
}
\vspace{1mm}
\noindent where $\left [ \boldsymbol{\cdot} \right ]_+$ denotes $max(0, \boldsymbol{\cdot})$,  $\mathcal{N}_{s,v}$ and $\mathcal{N}_{s,v'}$ are sets of randomly sampled negative points from $\mathcal{P}_{s, v}$ and $\mathcal{P}_{s,v'}$, respectively, and $m_p$ and $m_n$ are margins for positive and negative pairs, respectively. We set $|\mathcal{Z}|=64$ and $|\mathcal{N}|=10$. Following the values used by Xie et al. in \cite{xie2020pointcontrast}, we set the margins $m_p$ and $m_n$ to $0.1$ and $1.4$, respectively.

{\noindent\textit{2) Computing dense ``pseudo ground truth" heatmaps}}:
We found that training the \modelPointscore{} directly on the binary grasp labels in the \ourData{} results in poor generalization to test categories, as we show in our ablation studies in Section \ref{sec:sim_eval}. We hypothesize that this is because the model is susceptible to overfitting, due to the number of labeled points being a relatively small percentage of the total number of points in the input point cloud.
To mitigate this overfitting, we augment our data by assigning pseudo ground truth labels $h_{pgt}^{i}$ to each point $i$ in a point cloud with

{\small
\begin{align}\label{eq:heatmap}
    h_{pgt}^{i} &= \text{\scriptsize{min}} \left ( 
    1,
    \left (
        \sum\limits_{j \in G^{i}}
        \text{\scriptsize{max}}
        \left (
            0,w^{(i,j)}
        \right )
    \right )
    * \frac{1}{k}
    \right)\\
    w^{(i,j)} &=
    \begin{cases}
        1 - \frac{1}{r} * \lambda_+ d(i,j) & \text{if $j$ is positive}\\
        1 - \frac{1}{r} * \lambda_- d(i,j) & \text{if $j$ is negative},
    \end{cases}  
\end{align}
}

\noindent
where $h_{pgt}^{i}$ is a weighted average of the $k$ closest labeled points to point $i$, denoted by the set $G^i$. Each labeled point $j \in G^{i}$,  if within distance $r$ to point $i$, contributes a weight $w^{(i,j)}$ to the pseudo ground truth label of $i$ inversely proportional to its distance to $i$, where $\lambda_{\{+, -\}}$ is the amount to weight positive and negative points, and $d$ denotes the Euclidean distance between two points. Point $i$ is labeled as negative if the closest ground truth positive point is farther than $r$. We use $k=15$, $\lambda_+=2$, $\lambda_-=0$, and $r=4$cm. \mbox{Fig. \ref{fig:pipeline}b} illustrates the difference between raw binary labels and dense heatmaps.%

\noindent \textit{Total loss}: To train the \modelPointscore{}, we use the loss function
\vspace{0.5em}
\begin{dmath}\label{eq:total}
    \mathcal{L}_{\text{total}} = \lambda_{HC} \mathcal{L}_{HC} + \lambda_{MSE} \mathcal{L}_{MSE}
\end{dmath}, 
\vspace{0.5em}
which combines the hardest contrastive loss $\mathcal{L}_{HC}$ and the mean squared error $\mathcal{L}_{MSE}$ between per-point predicted scores and pseudo ground truth heatmap labels, weighted with $\lambda_{HC}$ and $\lambda_{MSE}$ respectively, to learn generalizable feature encodings. We set $\lambda_{HC}=3$ and $\lambda_{MSE}=1$.

\subsection{Predicting grasp orientations}

While considering an articulated object's actionability and joint configuration is critical for predicting good grasp points, these properties matter much less for predicting grasp orientations. Instead, if given a good grasp point, an object's local geometry is the most important factor in determining a suitable grasp orientation, regardless of whether the object is rigid or articulated. As such, we leverage orientation predictions from Contact-GraspNet (CGN) \cite{Sundermeyer-2021-CGN}, a state-of-the-art grasp generation method for rigid objects. As CGN is trained on point clouds with 2048 points, whereas our point clouds have 4096 points, we assign each point in our point cloud the orientation corresponding to the closest point in the down-sampled CGN point cloud. 

Although the \ourData{} contains 6 DoF grasps, the performance of the models we trained on our data to predict orientations could not match that of CGN's. We believe this is because of the difference in training data quantity and density between the \ourData{} and the ACRONYM dataset \cite{eppner2021acronym}, which CGN is trained on. The ACRONYM dataset is not only substantially larger than \ourData{}, containing over 17 million grasps, but its grasps are also on rigid objects much smaller in scale than the articulated objects in the \ourData{}, resulting in much denser coverage of ground truth grasps on objects. It remains exciting future work to develop less data-hungry methods for learning grasp orientations, as we have already done for learning grasp points.

\section{Experimental results}

\subsection{Simulation evaluation}
\label{sec:sim_eval}

In simulation, we compare \ourName{} to baseline methods and ablations. In the comparison against baseline methods, we investigate how state-of-the-art models for both rigid object grasping and interacting with articulated objects compare to \ourName{} on the task of grasping articulated objects. In our ablation study, we explore the effect that pre-training PointNet++ (PN++) and training on dense heatmap labels has on the \modelPointscore{} performance.

{\noindent\textbf{Evaluation setup}:} 
We train and test \ourName{} on partial point clouds captured from camera viewpoints sampled randomly from a range of 120\textdegree{} about the object's yaw axis and 10\textdegree{} about the object's pitch axis. We train on 47 instances (6 \texttt{Box}, 14 \texttt{Dishwasher}, 9 \texttt{Microwave}, 9 \texttt{Safe}, 9 \texttt{TrashCan}), with 8 states per instance (1 closed, 7 randomly-sampled open) and 16 randomly-sampled viewpoints per state. We test each model on 480 partial point clouds of 14 held-out training category instances (3 \texttt{Box}, 3 \texttt{Dishwasher}, 2 \texttt{Microwave}, 2 \texttt{Safe}, 4 \texttt{TrashCan}), and 920 partial point clouds of 23 held-out test category instances (6 \texttt{Oven} and 17 \texttt{StorageFurniture}). \ourName{} is only trained on objects with one movable part, but several test category instances have more than one movable part. For each partial point cloud, we evaluate the top-1 highest scoring grasps by executing them in simulation with the procedure described in Section \ref{sec:grasp-ep}. We report the success rates of the top-1 grasps for each test scene.

{\noindent\textbf{Baselines}:} \label{sec:ours-v-baselines}
To highlight why existing grasp generation methods for rigid objects are not suitable for grasping articulated objects, we compare \ourName{} to Contact-GraspNet (CGN) out-of-the-box, using both its proposed grasp points and orientations. We also compare \ourName{} to Where2Act (W2A) \cite{Mo-2021-W2A} and VAT-Mart \cite{Wu_2022-VATMART}, two existing methods that also consider interactions with articulated objects, but do not focus specifically on prehensile manipulation. We evaluate pre-trained VAT-Mart\footnote{VAT-Mart does not provide model checkpoints and the released code contains errors that prevent the user from training policies. After fixing these errors, training VAT-Mart, and running their evaluation, we verified that we were able to match the scores reported in their paper \cite{Wu_2022-VATMART}. This is the policy we use in our evaluation.\label{vatmart-fn}}
and W2A models using their `pull-door' and `pull' action policies, respectively. We evaluate each method both with and without ground truth segmentation masks.
First, to match how W2A and VAT-Mart conduct their evaluations\footnote{Although the use of segmentation masks may not be stated clearly in their papers, we found that the released W2A and VAT-Mart code use ground truth part segmentation masks during both training and evaluation.\label{segmentation-fn}}, 
we use ground truth part segmentation masks to filter out grasps on unmovable object parts.
Secondly, as our method does not need segmentation masks, we evaluate the proposals from W2A and VAT-Mart without masks.

\begin{table*}
\centering
\vspace{2.5mm}
\footnotesize
\setlength{\tabcolsep}{2pt} %
\begin{tabular}{cccccccccccccccccccc} 
\toprule
  & \multicolumn{19}{c} {\textit{Simulation success rates (\%) $\uparrow$}} 
  \\
  \cmidrule(lr){2-20}
 & \textbf{All} & \multicolumn{12}{c}{\textbf{Train categories}} & \multicolumn{6}{c}{\textbf{Test categories}} \\
\cmidrule(lr){2-2} \cmidrule(lr){3-14} \cmidrule(lr){15-20}
 & \textit{All states} & \multicolumn{6}{c}{\textit{Closed state}} & \multicolumn{6}{c}{\textit{Open states}} & \multicolumn{3}{c}{\textit{Closed state}} & \multicolumn{3}{c}{\textit{Open states}}\\
\cmidrule(lr){2-2} \cmidrule(lr){3-8} \cmidrule(lr){9-14} \cmidrule(lr){15-17} \cmidrule(lr){18-20}
 & \textbf{All} & \textbf{All} & \Huge{\cardbox} & \Huge{\dishwasher} & \Huge{\microwave} & \Huge{\safe} & \Huge{\trash} & \textbf{All} & \Huge{\cardbox} & \Huge{\dishwasher} & \Huge{\microwave} & \Huge{\safe} & \Huge{\trash} & \textbf{All} & \Huge{\oven} & \Huge{\stor} & \textbf{All} & \Huge{\oven} & \Huge{\stor} \\ 
 \textbf{Model} & n\,=\,1400 & 200 & 20 & 60 & 20 & 20 & 80 & 280 & 60 & 60 & 40 & 40 & 80 & 460 & 120 & 340 & 460 & 120 & 340 \\
\midrule
\ourName{} (Ours) & \textbf{45.0} & \textbf{40.0} & \textbf{65.0} & \textbf{48.3} & \textbf{35.0} & \textbf{10.0} & 36.3 & \textbf{56.1} & 55.0 & 55.0 & \textbf{85.0} & 42.5 & \textbf{50.0} & \textbf{37.4} & \textbf{45.0} & \textbf{34.7} & \textbf{48.0} & \textbf{29.2} & \textbf{54.7}\\
CGN~\cite{Sundermeyer-2021-CGN} & 35.0 & 29.0 & 30.0 & 0.0 & 30.0 & 0.0 & \textbf{57.5} & 46.8 & \textbf{58.3} & \textbf{56.7} & 57.5 & \textbf{50.0} & 23.8 & 21.7 & 7.50 & 26.8 & 43.7 & 27.5 & 49.4\\
VAT-Mart w/ mask \cite{Wu_2022-VATMART} & 5.21 & 3.0 & 0.0 & 10 & 0.0 & 0.0 & 0.0 & 6.8 & 6.7 & 1.7 & 7.5 & 2.5 & 12.5 & 0.6 & 2.5 & 0.0 & 9.8 & 15.8 & 7.6\\

VAT-Mart w/o mask ~\cite{Wu_2022-VATMART} & 2.0 & 1.0 & 0.0 & 3.3 & 0.0 & 0.0 & 0.0 & 5.4 & 3.3 & 6.7 & 2.5 & 0.0 & 10.0 & 0.0 & 0.0 & 0.0 & 2.4 & 3.3 & 2.1\\

W2A w/ mask ~\cite{Mo-2021-W2A} & 3.9 & 2.0 & 5.0 & 1.7 & 0.0 & 0.0 & 2.5 & 8.9 & 6.7 & 5.0 & 22.5 & 10.0 & 6.3 & 0.0 & 0.0 & 0.0 & 5.4 & 3.3 & 6.2\\

W2A w/o mask ~\cite{Mo-2021-W2A} & 2.4 & 0.5 & 0.0 & 1.7 & 0.0 & 0.0 & 0.0 & 4.3 & 0 & 0.0 & 10 & 10 & 5 & 0.0 & 0.0 & 0.0 & 4.4 & 0.0 & 5.9\\
\bottomrule
\end{tabular}
\caption[]{\label{tab:ours-v-baselines-small}Simulation grasp success rates (\%) for the top-1 grasps generated by \ourName{}, CGN, VAT-Mart \cite{Wu_2022-VATMART}, and Where2Act (W2A) \cite{Mo-2021-W2A}. Results are broken down by \modelPointscore{} train/test categories (all instances are unseen during training) as well as open and closed states. Note that pre-trained W2A and VAT-Mart have different train/test categories: W2A trains on \texttt{Box}, \texttt{Microwave}, \texttt{TrashCan}, and \texttt{StorageFurniture} and VAT-Mart trains on  \texttt{StorageFurniture} and \texttt{Microwave}. Given that VAT-Mart and W2A use part segmentation masks\footref{segmentation-fn} during training and evaluation, we include results for both with and without part segmentation. \ourName{} achieves a higher average success rate than all baselines. We note that the success rates for closed-state \texttt{Safe} instances are lower than success rates of other categories because the handles on them are very small and shallow, making them difficult to grasp. Even still, \ourName{} achieves a 10\% success rate, while all other baselines fail to find any successful grasps.}
\end{table*}

\begin{table*}
\centering
\footnotesize
\setlength{\tabcolsep}{3pt} %
\begin{tabular}{ccccccccccccccccccc} 
\toprule
& \multicolumn{18}{c} {\textit{Real-world success rates (\%) $\uparrow$}} 
  \\
  \cmidrule(lr){2-19}
 & \textit{All states} & \multicolumn{9}{c}{\textit{Closed states}} & \multicolumn{8}{c}{\textit{Random states}}\\
\cmidrule(lr){2-2}\cmidrule(lr){3-10} \cmidrule(lr){11-19}
 & \textbf{All} & \textbf{All} & \Huge{\handlecyl} & \Huge{\handleL} & \Huge{\handlesmooth} & \Huge{\handletex} & \Huge{\drawer} & \Huge{\microwave} & \Huge{\toaster} & \textbf{All} & \Huge{\hingeright} & \Huge{\hingeleft} & \Huge{\hingeup} & \Huge{\hingedown} & \Huge{\drawer} & \Huge{\microwave} & \Huge{\toaster} & \Huge{\cardbox} \\ 
 \textbf{Model} & n\,=\,120 & 56 & 8 & 8 & 8 & 8 & 8 & 8 & 8 & 64 & 8 & 8 & 8 & 8 & 8 & 8 & 8 & 8\\
\midrule
\ourName{} (Ours) & \textbf{67.5} & \textbf{57.1} & \textbf{87.5} & \textbf{62.5} & \textbf{75.0} & \textbf{37.5} & \textbf{37.5} & \textbf{62.5} & \textbf{37.5} & \textbf{76.6} & \textbf{100} & \textbf{100} & \textbf{37.5} & \textbf{87.5} & \textbf{62.5} & \textbf{87.5} & \textbf{50.0} & \textbf{87.5} \\
CGN~\cite{Sundermeyer-2021-CGN} & 33.3 & 10.7 & 0.00 & 12.5 & 0.00 & 0.00 & 0.00 & 25.0 & \textbf{37.5} & 53.1 & 62.5 & 50.0 & \textbf{37.5} & \textbf{87.5} & \textbf{62.5} & 62.5 & 0.00 & 62.5\\
\bottomrule
\end{tabular}
\caption{\label{tab:real-exp}Real-world grasp success rates (\%) for \ourName{} compared to baseline CGN. The first column of results lists the average success rates over all 120 scenes, where AO-Grasp achieves a success rate double that of CGN. The following columns break down results by joint state, reconfigurable cabinet variation, and object type. For closed-state reconfigurable cabinet variations, we break down results by handle type (cyndrical, L-shaped, smooth climbing hold, textured climbing hold), as local handle geometry is the most critical factor for grasping objects in closed states. For open-state reconfigurable cabinet variations, we break down results by articulation axes (hinge-right, hinge-left, hinge-up, hinge-down), as an object's articulation axis is more relevant than handle geometry when grasping objects in open states. Notably, \ourName{} dramatically outperforms CGN on closed states, where CGN fails to generate any successful grasps for 4 out of the 8 closed state object variations. Please see the supplementary video for more details on real-world experiments.}
\vspace{-5mm}
\end{table*}

\begin{figure}[h]
\centering
\includegraphics[width=\linewidth]{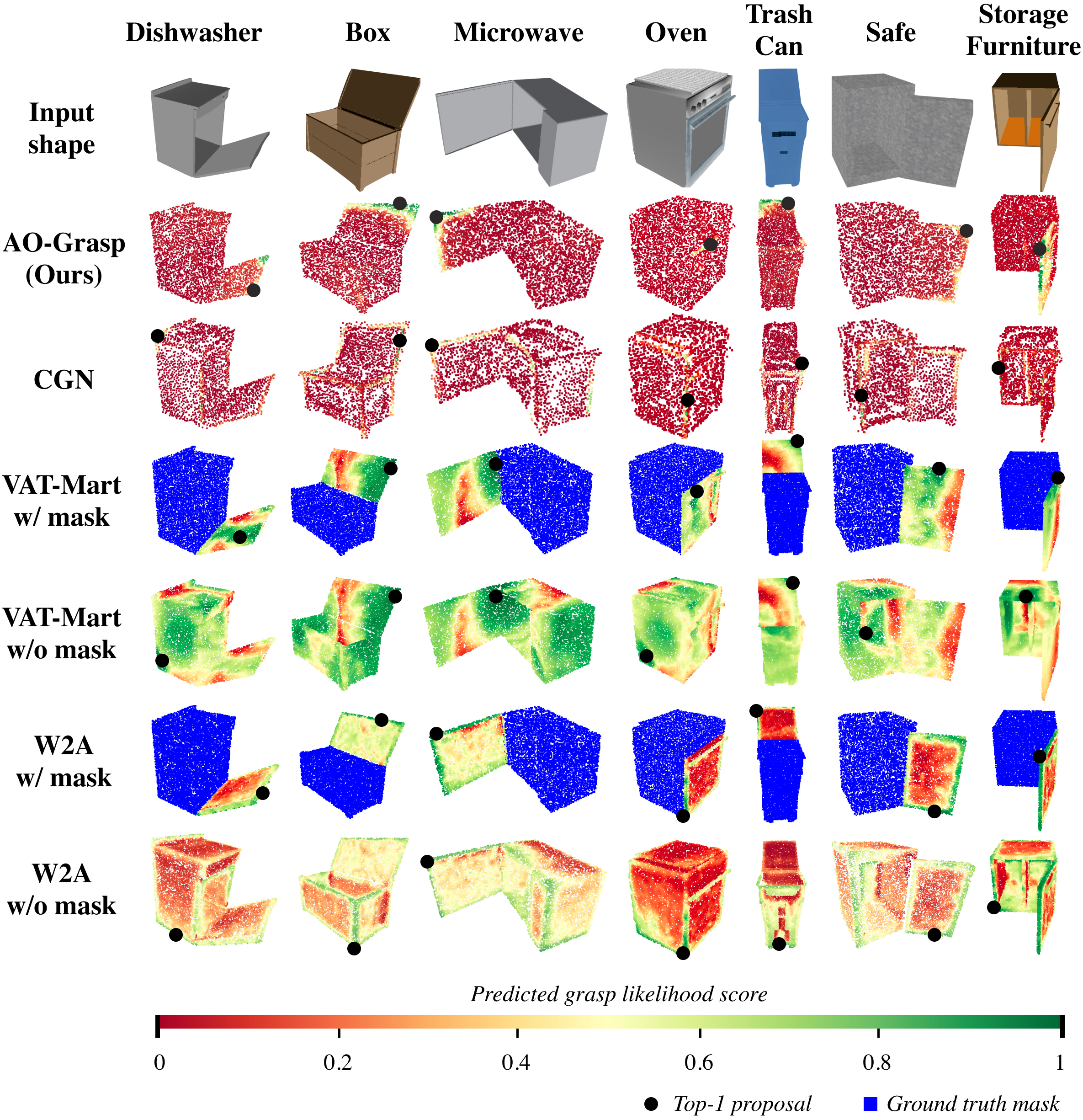}
\caption{A comparison of predicted grasp-likelihood scores from \ourName{} and baselines CGN, VAT-Mart, and W2A on synthetic point clouds, with \mbox{top-1} proposals highlighted with black dots. Given that VAT-Mart and W2A use part-segmentation masks\footref{segmentation-fn} in their training and evaluation, while \ourName{} does not, we include their predictions with and without ground truth masks. Compared to \ourName{}, all baselines tend to propose non-actionable points more often. Point cloud sizes are 4K, 2K, 10K, and 10K for \ourName{}, CGN, VAT-Mart, and W2A, respectively.}
\label{fig:predicted-heatmaps}
\vspace{-3mm}
\end{figure}

{\noindent\textbf{Results against baselines}:} 
We show in Table \ref{tab:ours-v-baselines-small} that \ourName{} consistently achieves higher success rates than CGN, W2A, and VAT-Mart on both closed and open states. This trend holds true even on categories not seen by \ourName{} during training, demonstrating that \ourName{} achieves generalization across categories, even to objects that contain multiple joints. Compared to CGN, which often proposes grasps on unmovable object parts, \ourName{} predicts grasp scores that capture actionability and thus proposes grasps that are on the movable parts of objects, as shown in Fig. \ref{fig:predicted-heatmaps}. While CGN suffers from this failure mode in all states, closed states, particularly of the test categories, emphasize this more as the number of graspable points in closed states is much smaller compared to open states (i.e. the points on the handle are much fewer than those on the edges of a door).

Without using ground truth masks, W2A, VAT-Mart exhibit similar failure cases to CGN, where grasps are often on unmovable object parts (Fig. \ref{fig:predicted-heatmaps}). However, even when we use ground truth segmentation masks to ensure that grasps are on movable object parts, both methods still achieve very low success rates, as the proposed grasp orientations do not lead to stable grasps. In general, unlike CGN, W2A and VAT-Mart hardly find any successful grasps because they are not trained specifically for grasping and thus struggle to perform well under our grasp evaluation criteria. Compared to W2A, VAT-Mart achieves a slightly higher success rate, likely because it considers task-aware pulling trajectories which improves the quality of the gripper orientation predictions.

\begin{figure}[h]
\centering
\includegraphics[width=\linewidth]{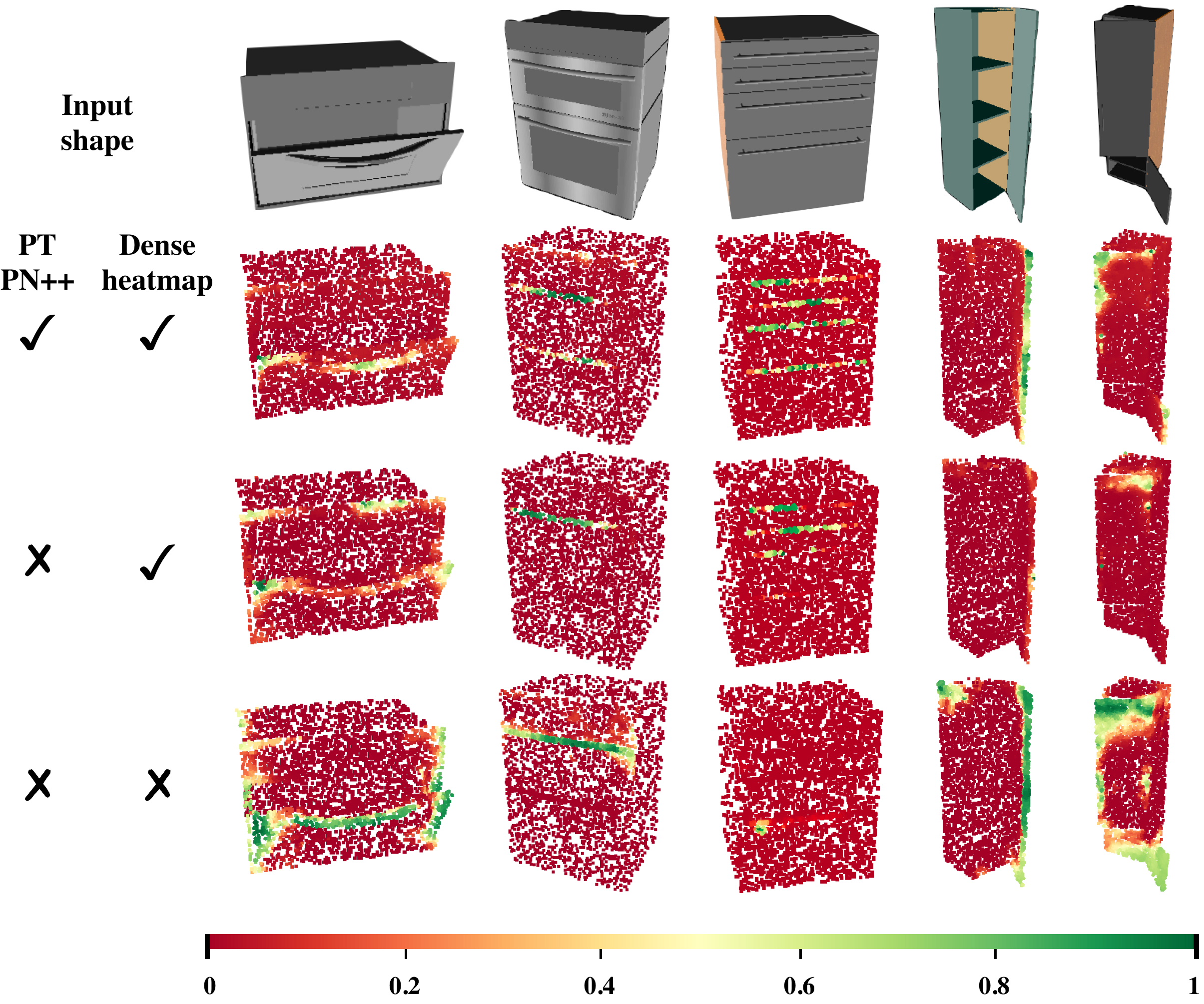}
\caption{\label{fig:ablation-heatmaps}A comparison of predicted grasp-likelihood scores from \ourName{} and ablations on synthetic point clouds of test category instances. Despite not being trained on any instances with more than one movable part, \ourName{} (first row of heatmaps) accurately predicts scores for objects with multiple movable parts. For example, in the second and third columns, it detects points on all the handles, and in the last column it detects points on the edges of both doors. Moreover, it rarely predicts false positives, i.e., it does not predict points on the body of the object. In contrast, the model without pre-training PN++ and dense heatmaps (third row of heatmaps) does not predict grasp points on all handles (second and third columns) and predicts many false positives. Training the model with dense heatmaps (second row of heatmaps) reduces the number of false positives (i.e. top left corner of the cabinet in the fourth column). However, the number of false negatives (missed good points) also increases (ie. bottom door of the right-most cabinet).}
\label{fig:ablation-heatmaps}
\vspace{-7mm}
\end{figure}

{\noindent\textbf{Ablations}:}  
\label{sec:ours-ablation}
In our ablation studies, we explore how pre-training PN++ on viewpoint-independent point correspondences and training the \modelPointscore{} on dense heatmap labels influence model performance. Table \ref{tab:ablation-success} shows the grasp success rates for the top-1 proposals generated by an \modelPointscore{} with a pre-trained and finetuned PN++ and trained on dense heatmaps (our method), as well as 2 ablation models: one without pre-trained PN++ but still trained with dense heatmaps and one with neither pre-trained PN++ nor dense heatmaps. Overall, the model with pre-trained PN++ and trained on dense heatmaps outperforms both ablation models. Further examination of the train and test category results reveals that both pre-training PN++ and supervising on dense heatmap labels improve generalization to test categories. Notably, while the model trained with sparse binary labels is competitive with our full method on train categories, it performs worse on test categories, particularly on closed states. This difference in performance underscores the role that our pseudo ground truth heatmaps play in mitigating model overfitting.

Additionally, in Table \ref{tab:ablation-f-score}, we report the F-score and accuracy rates for the predicted grasp-likelihood scores. These results show that pre-training PN++ and training on dense heatmap labels improves the model's ability to capture graspable regions for both train and test categories. In Figure \ref{fig:ablation-heatmaps}, which shows several examples of predicted \scoreName{} scores from each model on test category instances, we see that both ablation models are more prone to predicting false positives and false negatives.

\begin{table}[h]
\centering
\footnotesize
\setlength{\tabcolsep}{3pt} %
\begin{tabular}{ccccccc} 
\toprule
& & \multicolumn{5}{c}{\textit{Ablations' success rate (\%) $\uparrow$}}\\
\cmidrule(lr){3-7}
 & & \textbf{All} & \multicolumn{2}{c}{\textbf{Train categories}} & \multicolumn{2}{c}{\textbf{Test categories}} \\
\cmidrule(lr){3-3}\cmidrule(lr){4-5}\cmidrule(lr){6-7}
 & & \textit{All states} & \textit{Closed} & \textit{Open} & \textit{Closed} & \textit{Open}\\
\cmidrule(lr){3-3}\cmidrule(lr){4-4}\cmidrule(lr){5-5}\cmidrule(lr){6-6}\cmidrule(lr){7-7}
\textbf{PT} & \textbf{Dense} & n\,= & n\,= & n\,= & n\,= & n\,= \\ 
\textbf{PN++} & \textbf{heatmap} & 1400 & 200  & 280 & 460 & 460 \\
\midrule
\cmark & \cmark & \textbf{45.0} & 40.0 & 56.1 & \textbf{37.4} & \textbf{48.0}\\
\xmark & \cmark & 41.7 & 34.5 & 59.3 & 28.3 & 47.4\\
\xmark & \xmark & 42.2 & \textbf{41.5} & \textbf{60.4} & 33.9 & 39.8\\
\bottomrule
\end{tabular}
\caption{\label{tab:ablation-success}Simulation top-1 grasp success rates for \modelPointscore{} ablations on pre-training PointNet++ (PT PN++) and training on dense pseudo ground truth heatmaps. Without PT PN++ (second row) and without supervising on dense heatmaps (third row), generalization to test categories suffers, particularly on closed states for test categories.}
\vspace{-5mm}
\end{table}

\begin{table}[h]
\centering
\footnotesize
\setlength{\tabcolsep}{1.8pt} %
\begin{tabular}{ccccccc} 
\toprule
 & & \multicolumn{5}{c}{\textit{F-score (\%) $\uparrow$ \, / \, Accuracy (\%) $\uparrow$}}\\
 \cmidrule(lr){3-7}
 & & \textbf{All} & \multicolumn{2}{c}{\textbf{Train categories}} & \multicolumn{2}{c}{\textbf{Test categories}} \\
\cmidrule(lr){3-3}\cmidrule(lr){4-5}\cmidrule(lr){6-7}
 & & \textit{All states} & \textit{Closed} & \textit{Open} & \textit{Closed} & \textit{Open}\\
\cmidrule(lr){3-3}\cmidrule(lr){4-4}\cmidrule(lr){5-5}\cmidrule(lr){6-6}\cmidrule(lr){7-7}
\textbf{PT} & \textbf{Dense} & n\,= & n\,= & n\,= & n\,= & n\,= \\ 
\textbf{PN++} & \textbf{heatmap} & 1400 & 200  & 280 & 460 & 460 \\
\midrule
\cmark & \cmark &  \textbf{36.4}\,/\,\textbf{73.1} &   \textbf{45.4}\,/\,\textbf{71.6} &     40.3\,/\,\textbf{85.7} &  \textbf{54.1}\,/\,\textbf{72.8} & \textbf{15.0}\,/\,\textbf{62.0}\\
\xmark & \cmark &  33.5\,/\,67.4 &   36.8\,/\,68.3 &  \textbf{41.0}\,/\,80.0 &  43.5\,/\,66.5 & 11.3\,/\,56.1\\
\xmark & \xmark &  2.80\,/\,50.5 &   2.42\,/\,50.5 &     3.24\,/\,50.4 &  3.32\,/\,50.4 & 2.00\,/\,50.2\\
\bottomrule
\end{tabular}
\caption{\label{tab:ablation-f-score} F-score and accuracy rates of predicted \scoreName{} scores on synthetic point clouds for \modelPointscore{} ablations on pre-training PointNet++ (PT PN++) and training on our dense pseudo ground truth heatmaps. Without PT PN++ (second row), and without supervising on dense heatmaps (third row), models achieve lower F-scores, indicating a higher rate of false positives and false negatives. For all models, the F-score for open-state test categories is lower than F-scores for other splits. This is because the test categories have objects with multiple movable parts and therefore more graspable areas, resulting in the sampled grasps in the \ourData{} that are more spread out.
Upon visualizing the predicted \scoreName{} scores (Fig. \ref{fig:ablation-heatmaps}), we find that the lower F-scores are because the models are able to capture good graspable regions that are not sampled in the \ourData{}.}
\vspace{-5mm}
\end{table}

\subsection{Real-world evaluation} \label{sec:real-data}

To showcase \ourName{}'s zero-shot sim-to-real transfer, we conduct a quantitative evaluation of AO-Grasp and CGN on 120 scenes of real-world objects with varied local geometries and articulation axes, in different joints states, and captured from different viewpoints. To comprehensively test different local geometries and articulation axes, we design and fabricate a custom reconfigurable cabinet. It features a magnetic handle mounting system that allows for easily exchanging custom door handles and can be flipped on any side to achieve different articulation axes. The top row of Fig. \ref{fig:teaser}c shows variations of the reconfigurable cabinet with climbing holds and 3D-printed cylindrical handles, as well as in both hinge-right and hinge-up configurations.

We evaluate both methods on 16 variations of our reconfigurable cabinet (4 handles, with 4 locations of the articulation axes per handle (hinge top, bottom, left, right)), as well as 2 microwaves, a prismatic-jointed drawer, toaster oven, and cardboard box. We test each object in both closed and open states, and viewed from multiple viewpoints. Note that for all of these objects, the models only receive partial point clouds of the object, which we obtain using \cite{kirillov2023segment}, as input; no additional information such as the type of articulation mechanism is provided. We use a Franka Emika robot arm and a ZED2 camera to capture depth data. We evaluate the top-1 grasp proposed by each model by moving the end-effector to that pose, closing the gripper, and executing an action to actuate grasped part for 3 seconds.

To actuate the part, we use a compliant controller to move the end-effector along a direction of motion, parameterized as a vector defined in the local object part frame. Because \ourName{}'s prehensile grasps result in the end-effector being rigidly attached to the object part, we can define the object part frame to be the end-effector frame and directly specify the direction of motion in the end-effector frame, removing the need to track the object part pose at every timestep. The compliant controller moves the end-effector along the local direction of motion while being compliant in all other directions. This approach allows us to use same vector to parameterize a controller for opening a revolute door and a controller for opening a prismatic drawer, instead of needing to plan different trajectories for each joint type. In our experiments, we define the direction of motion to be normal to the object part surface, computed from the object's point cloud, but we note that with the prehensile grasps proposed by \ourName{}, a robot could also move its end-effector in any direction without losing contact with the object.

We use the same success criteria that we used in our data generation and simulation evaluation, where a grasp is labeled a success if the target joint of the object is actuated and the end-effector maintains contact with the object for the entire interaction. The results, presented in Table \ref{tab:real-exp}, show that AO-Grasp achieves an overall success rate of 67.5\% while CGN achieves a success rate of 33.3\%. Our real-world success rate is higher than our simulation success rate because the grasp execution procedure used in simulation is less forgiving than the procedure used for real-world experiments. In simulation, we initialize the end-effector directly at grasp poses and label grasps as failures if the end-effector intersects the object by any amount. In the real-world, we use a compliant controller to move the end-effector to grasp poses. This is more forgiving than spawning the end-effector at grasp poses and checking for collision, resulting in fewer failed grasps in the real-world. Please see \href{https://stanford-iprl-lab.github.io/ao-grasp/}{the project website} to view some real-world grasps.

\section{Conclusion}

We introduce \ourName{}, which generates stable and actionable 6 DoF grasps on articulated objects, and the \ourData{}, which contains 78K simulated grasps. Although \ourName{} achieves higher grasp success rates than baselines and shows promising sim-to-real transfer on a variety of objects, we acknowledge that there is still much room for improvement in both object diversity and model performance, underscoring the difficulty of grasp generation for articulated objects and leaving room for future work.

\newpage
\bibliographystyle{IEEEtran}
\bibliography{references}

\end{document}